\documentclass[sn-nature]{sn-jnl}

\usepackage{graphicx, subfigure}%
\usepackage{multirow}%
\usepackage{amsmath,amssymb,amsfonts}%
\usepackage{amsthm}%
\usepackage{mathrsfs}%
\usepackage[title]{appendix}%
\usepackage{xcolor}%
\usepackage{textcomp}%
\usepackage{manyfoot}%
\usepackage{booktabs}%
\usepackage{algorithm}%
\usepackage{algorithmicx}%
\usepackage{algpseudocode}%
\usepackage{listings}%

\usepackage{hyperref}
\makeatletter
\def\UrlAlphabet{%
      \do\a\do\b\do\c\do\d\do\e\do\f\do\g\do\h\do\i\do\j%
      \do\k\do\l\do\m\do\n\do\o\do\p\do\q\do\r\do\s\do\t%
      \do\u\do\v\do\w\do\x\do\y\do\z\do\A\do\B\do\C\do\D%
      \do\E\do\F\do\G\do\H\do\I\do\J\do\K\do\L\do\M\do\N%
      \do\O\do\P\do\Q\do\R\do\S\do\T\do\U\do\V\do\W\do\X%
      \do\Y\do\Z}
\def\UrlDigits{\do\1\do\2\do\3\do\4\do\5\do\6\do\7\do\8\do\9\do\0}
\g@addto@macro{\UrlBreaks}{\UrlOrds}
\g@addto@macro{\UrlBreaks}{\UrlAlphabet}
\g@addto@macro{\UrlBreaks}{\UrlDigits}
\makeatother

\usepackage{caption}

\newtheorem{theorem}{Theorem}
\newtheorem{proposition}[theorem]{Proposition}%

\begin{document}

\title[OT-Net: A Reusable Neural Optimal Transport Solver]{OT-Net: A Reusable Neural Optimal Transport Solver}


\author[1,4]{\sur{Zezeng Li}}\email{zezeng.lee@gmail.com}
\equalcont{These authors contributed equally to this work.}

\author[1]{\sur{Shenghao Li}}\email{lshymfl@126.com}
\equalcont{These authors contributed equally to this work.}

\author[2]{\sur{Lianbao Jin}}\email{1099630577@qq.com}

\author*[3]{\sur{Na Lei}}\email{nalei@dlut.edu.cn}
\author[1]{\sur{Zhongxuan Luo}}\email{zxluo@dlut.edu.cn}

\affil[1]{\orgdiv{School of Software}, \orgname{Dalian University of Technology}, \orgaddress{\city{Dalian}, \postcode{116620}, \country{China}}}

\affil[2]{\orgdiv{School of Mathematical Sciences}, \orgname{Dalian University of Technology}, \orgaddress{ \city{Dalian}, \postcode{116024}, \country{China}}}

\affil[3]{\orgdiv{ DUT-RU ISE}, \orgname{Dalian University of Technology}, \orgaddress{\city{Dalian}, \postcode{116620}, \country{China}}}

\affil[4]{\orgdiv{Beijing Key Laboratory of Light-field Imaging and Digital Geometry}, \orgname{Capital Normal University}, \orgaddress{\city{Beijing}, \postcode{100048}, \country{China}}}

\abstract{With the widespread application of optimal transport (OT), its calculation becomes essential, and various algorithms have emerged. However, the existing methods either have low efficiency or cannot represent discontinuous maps. A novel reusable neural OT solver \textbf{OT-Net} is thus presented, which first learns Brenier's height representation via the neural network to get its potential, and then obtains the OT map by the gradient of the potential. The algorithm has two merits: 1) When new target samples are added, the OT map can be calculated straightly, which greatly improves the efficiency and reusability of the map. 2) It can easily represent discontinuous maps, which allows it to match any target distribution with discontinuous supports and achieve sharp boundaries, and thus eliminate mode collapse.
Moreover, we conducted error analyses on the proposed algorithm and demonstrated the empirical success of our approach in image generation, color transfer, and domain adaptation. }




\maketitle

\section{Introduction}
The OT map transmits one probability measure to another in the most economical way,  which is widely applied in various areas, such as 
generative model \cite{seguy2018,chen2019gradual,An2020AEOT,liu2019wganqc,Daniels2021,Rout2021,gulrajani2017improvedgan}, domain adaptation \cite{courty2017joint,damodaran2018deepjdot,wang2021,chang2022unified,rakotomamonjy2022optimal,chuang2023infoot,tran2023unbalanced}, color transfer \cite{strossner2023low,bonneel2019spot,Alvarez2018,bonneel2016wasserstein,Ferradans2014}, up-sampling \cite{li2022real,Gazdieva2022,li2022weakly,golla2020temporal}. 
Consequently, an effective algorithm to solve the OT problem is essential. It can be regarded as a standard linear programming (LP) task, which can be computed by LP tools. However, in practical issues, the dimension of two distributions is usually thousands upon thousands. In this situation, the computational complexity is unaffordable if the problem is solved by LP algorithms. Later, Cuturi et al. \cite{cuturi2013sinkhorn} propose the Sinkhorn-Knopp algorithm by adding an entropic regularizer into the original OT problem, which can solve it quickly by sacrificing accuracy. Therefore, various improved variants have emerged, such as the iterative Bregman projections algorithm \cite{bjd2015}, the adaptive primal-dual accelerated algorithm \cite{dpa2018}, the inexact proximal point method \cite{xie2020}, and an accurate algorithm based on Nesterov's smoothing technique \cite{anaaai2022}.

Although these algorithms can effectively address the OT problem, learning or even approximating such an OT plan is computationally challenging for large and high-dimensional datasets due to the intrinsic curse of dimensionality. Thus, some academics have proposed using neural networks to compute OT map  \cite{seguy2018,Makkuva2020,Daniels2021,Fan2021,Korotin2022,Gazdieva2022,Rout2021,Asadulaev2022} to mitigate these problems. Among them, Seguy et al. \cite{seguy2018} propose a two-step approach. First, they employ a simple dual stochastic gradient algorithm for solving regularized OT, then they estimate a Monge map as a neural network learned by approximating the barycentric projection of the OT plan. This method requires two optimization processes, which directly reduces the efficiency of this algorithm. 
Makkuva et al.~\cite{Makkuva2020} present a new framework to estimate the OT map based on input convex neural networks. Although this algorithm improves the model's performance, 
it needs to calculate two potential functions simultaneously, which directly doubles the size of parameters and floating-point operations. 

Other researchers~\cite{An2020AEOT,lei2019geometric,gu2016minkowski} attempt to provide OT-solving algorithms from a convex geometric perspective. When the cost function is quadratic, Brenier theorem \cite{Brenier1991} indicates that the OT map is given via the gradient of a piece-wise convex function which is called Brenier's potential. As~\cite{An2020AEOT,lei2019geometric,gu2016minkowski} remark, Brenier's potential is obtained by optimizing a convex energy, its solution is globally optimal. This means that the OT map has the same characteristics. Whereas, when the number of target domain data changes, AE-OT~\cite{An2020AEOT} has to re-optimize the convex energy equation to compute the OT map, resulting in low reusability. \textcolor{blue}{Imagine that we need to generate a specific facial image that has not been seen during the training process, perform interpolation or other editing on it, or generate animated characters that do not exist in the training dataset. At this point, we definitely don't want to resolve the OT map on the entire dataset. Our preferred approach is to directly add these new data to the calculation of the OT map.}

Consequently, this paper presents a novel neural network-based method to work out the above problem. Note that our algorithm is different from other neural network-based methods \cite{seguy2018,Daniels2021,Rout2021}. While our algorithm learns Brenier's potential via a single neural network, and the OT map is obtained by computing its gradient which endows our algorithm with the ability to represent discontinuous maps, this can effectively eliminate mode collapse during the generation process. Moreover, compared to AE-OT~\cite{An2020AEOT}, our method has higher reusability. When new target samples are added, AE-OT needs to optimize and solve the OT problem from scratch, but the proposed algorithm can calculate the OT map straightforwardly via the learned height representation without retraining or re-optimization.

In summary, our main \textbf{contributions} are as follows:
\begin{itemize}
    \item We propose a novel algorithm to compute the OT map, so-called the reusable neural optimal transport solver. Our algorithm can compute the OT map straightly when adding new target samples without retraining or re-optimization, which significantly improves its computational efficiency and reusability. 
    \item The algorithm enables the representation of discontinuous maps, which could perfectly avoid mode collapse in the generation.
    \item We theoretically analyze the error bound of the height vector, and the experimental results show that the algorithm has comparable performance in generating models, color transfer, and domain adaptation.
\end{itemize}

\section{Background on optimal transport}
\label{sec:background}
Optimal transport can be traced back to the seminal work of Monge~\cite{monge1781memoire}, in which the profound and far-reaching Monge's problem was raised.

\noindent\textbf{Monge’s problem}: Let $\mu  \sim {\cal P}(X)$and $\nu  \sim {\cal P}(Y)$ be two sets of  probability measures defined on $X$ and $Y$, respectively. Let cost function $c(x,y):X \times Y \to [0, + \infty ]$ measure the cost of transporting $x \in X$ to $y \in Y$. The Monge's problem seeks the most efficient $\mu$-measurable map $T:X \to Y$ by 
\begin{equation}
	\begin{array}{c}
		\mathop {\inf }\limits_T \int_X {c\left( {x,T(x)} \right)} d\mu (x)\\
		{\rm{subject \ to\ }}~~\nu  = {T_\# }\mu ,
	\end{array}
	\label{eq:Monge}
\end{equation}
where ${T_\# }\mu$ is the push-forward measure induced by $T$. A minimum ${T^*}$ to this problem is called an OT map. Intuitively, Monge’s problem finds a transport to turn the mass of $\mu $ into $\nu $ at the minimal cost measured by the cost function $c$. However, it has two drawbacks, $\mu$-mass cannot be split leading to hard constraint; its transport map may not exist.


To overcome the above shortcomings, Kantorovich \cite{kantorovich1942transfer} relaxed  transport maps into transport plans,
and in turn, raised the Kantorovich problem. Later, to speed up the computation of OT, the \textbf{regularized OT} is achieved by adding a negative-entropy penalty to the Kantorovich problem. One of the most representative entropy-regularization-based algorithms is the Sinkhorn-Knopp algorithm \cite{cuturi2013sinkhorn}. Although it reduces the computational complexity of the OT problem, the algorithm does not scale well to measures supported on a large number of samples, since each of its iterations has an $\mathcal{O}(n^2)$ complexity. Various improved algorithms thus have emerged to eliminate the above problems. However, due to an intrinsic curse of dimensionality, learning or even approximating such a map is computationally challenging for large and high-dimensional datasets.

Recently, \textbf{neural-network-based} OT solvers for high-dimensional settings have emerged, which are mainly divided into two ways. The first is computing the \textbf{OT cost} and using it as the loss function \cite{gulrajani2017improvedgan,liu2019wganqc,li2022weakly,Petzka2017,Sanjabi2018}. The second is the \textbf{OT map} itself can be used as a generative model \cite{Makkuva2020,Daniels2021, Rout2021}. Specifically, Daniels et.al \cite{Daniels2021} proposes a method for solving entropy-regularized OT using neural networks, but it is extremely time-consuming to generate samples via the Langevin dynamics. Makkuva et.al~\cite{Makkuva2020} employ input convex neural networks to parametrize potentials in the dual problem and get the OT map by the gradient of potential. Rout et al. \cite{Rout2021} apply OT directly in ambient spaces, such as spaces of high-dimensional images.

According to the OT theory proposed by Chen and
Figalli~\cite{chen2017}, the OT map is discontinuous when the support of the target domain is non-convex. Nevertheless, neural networks can only express continuous mapping. Transport maps thus learned by the above algorithms \cite{Daniels2021, Rout2021} are continuous, while OT maps are discontinuous at singular sets \cite{lei2019geometric}, and this intrinsic conflict leads to mode collapse. From the perspective of convex geometry, Lei et al. \cite{An2020AEOT} propose a generative model called AE-OT which perfectly avoids mode collapse/mixture. In the following, we provide a brief description of the convex geometry-based OT algorithm.

\textbf{Convex geometry-based OT.} Brenier \cite{Brenier1991} discovered the intrinsic connection between the OT map and convex geometry, and gave the following theorem.
\begin{theorem}\label{thm:brenier}
\cite{brenier1991polar}
Suppose $\mu$ and $\nu$ are two probability measures
defined on $\boldsymbol{X}\subset \mathbb{R}^n$ and $\boldsymbol{Y}\subset\mathbb{R}^n$, respectively, and the transportation cost is the quadratic Euclidean
distance $c(\boldsymbol{x}, \boldsymbol{y})=\|\boldsymbol{x}-\boldsymbol{y}\|^{2}$. If $\mu$ is absolutely continuous and $\mu$ and $\nu$ have finite second-order moments, then there exists a convex function $u$:$\boldsymbol{X}\rightarrow \mathbb{R}$, such that the gradient map $\nabla u$ gives the unique solution to the Monge's problem, where u is called Brenier's potential, $\nabla u$ is called the Brenier map or the optimal mass transportation map. In general, $u$ is not unique. 
\end{theorem}

As the Theorem \ref{thm:brenier} remarks, the OT map is given by the gradient map of Brenier's potential $u$ which can be parametrized by a height vector $\boldsymbol{h}$. Alexander~\cite{alexandrov2005convex} provided the existence of the solution to the OT problem based on algebraic topology,  which is not constructive. Afterward, to solve the OT map, Gu et al.~\cite{gu2016minkowski} provided constructive proof based on the variational principle. It is described in detail as follows. 

Suppose the source measure $\mu$ defined on a convex domain $\Omega \subset \mathbb{R}^{d}$, the target domain is a discrete set, $\boldsymbol{Y}=\left\{\boldsymbol{y}_{1}, \boldsymbol{y}_{2}, \cdots, \boldsymbol{y}_{n}\right\}, \boldsymbol{y}_{i} \in \mathbb{R}^{d}$. The target measure is a Dirac measure $\nu=\sum_{i=1}^{n} \nu_{i} \delta\left(\boldsymbol{y}-\boldsymbol{y}_{i}\right), i=1,2, \ldots, n$, 
with the equal total mass as the source measure, $\mu(\Omega)=\sum_{i=1}^{n} \nu_{i}$. A cell decomposition is induced $\Omega=\bigcup_{i=1}^{n} W_{i}$ under OT map $T: \Omega \rightarrow \boldsymbol{Y}$, such that every $\boldsymbol{x}$ in each cell $W_{i}$ is mapped to the target $\boldsymbol{y}_{i}$, $T: \boldsymbol{x} \in W_{i} \mapsto \boldsymbol{y}_{i}$. If the $\mu$-volume of each cell $W_{i}$ equals to the $\nu$-measure of target domain, \textit{i.e.}, $T\left(W_{i}\right)=\boldsymbol{y}_{i}$, $\mu\left(W_{i}\right)=\nu_{i}$, The OT map $T$ is measure preserving, denoted as $T_{\#} \mu=\nu$.
The cost function is given by $c: \Omega \times \boldsymbol{Y} \rightarrow \mathbb{R}$, where $c(\boldsymbol{x}, \boldsymbol{y})$ represents the cost for transporting a unit mass from $\boldsymbol{x}$ to $\boldsymbol{y}$. The OT map is obtained by minimizing the total transport cost as follows,
\begin{equation}
T^{\ast}:=\arg \min _{T_{\#} \mu=\nu} \int_{\Omega} c(\boldsymbol{x}, T(\boldsymbol{x})) d \mu(\boldsymbol{x})=\arg \min _{T_{\#} \mu=\nu} \sum_{i=1}^{n} \int_{W_{i}} c\left(\boldsymbol{x}, \boldsymbol{y}_{i}\right) d \mu(\boldsymbol{x}).\label{eq:SDOT}
\end{equation}

Under this convex geometric setting, to calculate the OT map, Gu et al.~\cite{gu2016minkowski} reformulate Brenier's potential $u$ as $u_{\boldsymbol{h}}$ with the height vector $\boldsymbol{h}$, which is 

\begin{equation}
u_{\boldsymbol{h}} = \max_{i=1}^{n}\{\pi_{\boldsymbol{h},i}(\boldsymbol{x})\}=\max_{i=1}^{n}\{ \boldsymbol{x}^T \boldsymbol{y}_i + h_i\}, u_{\boldsymbol{h}}: \Omega \rightarrow \mathbb{R},\label{eq:uh}
\end{equation}
where $\pi_{\boldsymbol{h},i}(\boldsymbol{x}) =  \boldsymbol{x}^T\boldsymbol{y}_i + h_i$ is the hyperplane corresponding to $\boldsymbol{y}_i$, $\boldsymbol{h} = (h_1, h_2, ..., h_n)$ is the height vector, $h_i$ denotes the height of the $i$-th sample.

Based on the above elaboration, Brenier's potential $u_{\boldsymbol{h}}$ is deterministic by $\boldsymbol{h}$. As a result, given the target measure $\nu$, there exists potential in Eq.\eqref{eq:uh} whose projected volume of each support plane is equal to the given target measure $\nu_i$. The projection of the graph of $u_{\boldsymbol{h}}$ decomposes $\Omega$ into cells $W_{i}(h_i)$, each cell is the projection of the supporting plane $\pi_{\boldsymbol{h}, i}(\boldsymbol{x})$. That is, the key to obtaining the potential energy function is to optimize the height vector $\boldsymbol{h}$. Specifically, under the condition $\sum_i h_{i}=0$, the height vector $\boldsymbol{h}$ is the minimum argument of the following convex energy:
\begin{equation}
E(\boldsymbol{h})=\int_{0}^{\boldsymbol{h}} \sum_{i=1}^{n} w_{i}(\eta) \mathrm{d} \eta_{i}-\sum_{i=1}^{n} h_{i} \nu_{i} \ ,
\label{eq:energy}
\end{equation}
where $w_{i}(\eta)$ is the $\mu$-volume of $W_{i}(\eta)$. The gradient of Eq.~\eqref{eq:energy} is given by the following:
\begin{equation}
\nabla E(\boldsymbol{h})=\left[w_{1}(\boldsymbol{h})-\nu_{1}, w_{2}(\boldsymbol{h})-\nu_{2}, \ldots, w_{n}(\boldsymbol{h})-\nu_{n}\right]^{\mathrm{T}}.
\label{eq:Ehgradient}
\end{equation}
Hence, the convex energy $E(\boldsymbol{h})$ can be optimized simply by the gradient descent method. Yet this algorithm needs to recalculate or reoptimize the OT map in the case of changes in target domain samples $\boldsymbol{Y}$, resulting in low reusability. Consequently, this paper presents a novel neural-network-based algorithm to compute OT map, \textit{i.e.}, the reusable neural OT solver. Compared with the AE-OT, we provide a more efficient method to calculate the height vector and Brenier's potential.
\section{The Proposed Algorithm}\label{sec3}

\begin{figure}[htb!]
    \centering
    \includegraphics[width=0.85\textwidth]{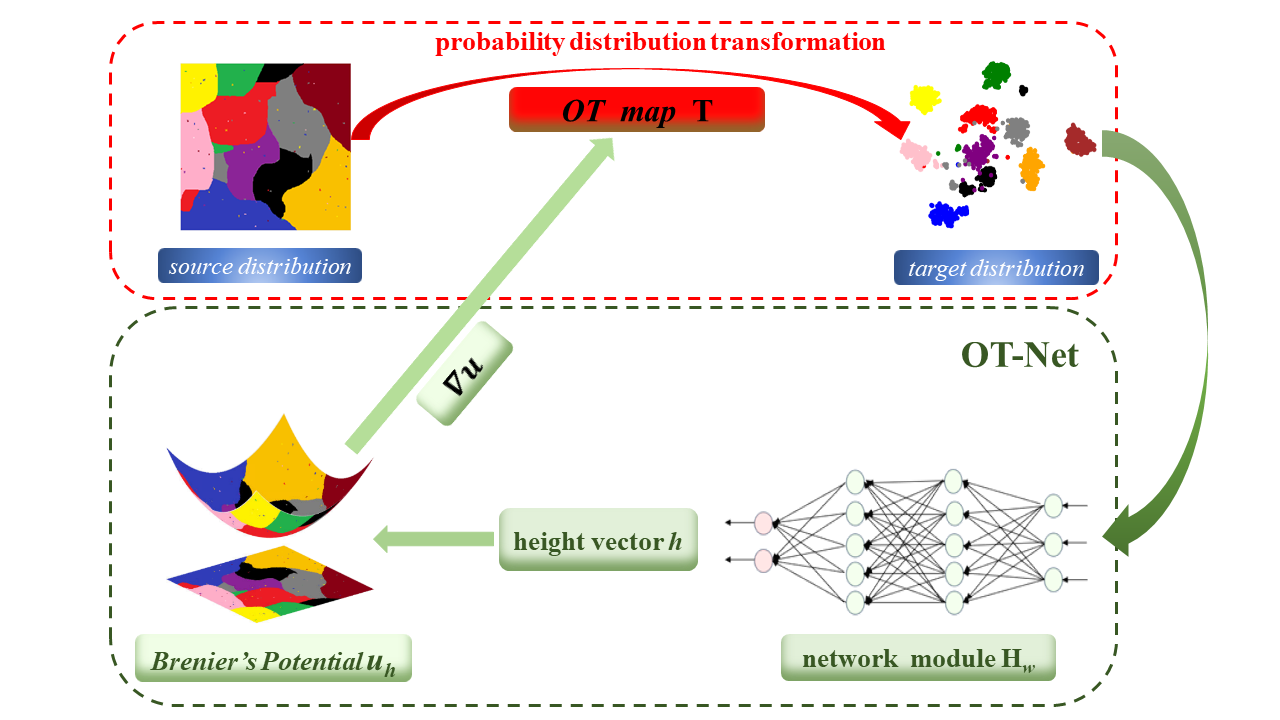}
    \caption{Architecture of our algorithm. \textbf{OT-Net} (green dotted box): First, the height vector $\boldsymbol{h}$ is obtained by a well-trained Brenier's height representation $H_{\Phi}$. Then, Brenier's potential $u_{\boldsymbol{h}}$ is calculated by Eq. \eqref{eq:ourbrenier}. Finally, the OT map is induced by its gradient. The \textbf{probability distributions transformation} (red dotted box): learning transformation of source distribution to target distribution via the OT map $T$.}
    \label{figure:OTnet} 
\end{figure}

The core task of deep learning is to learn the manifold structure of data and transform probability distributions. OT has a natural advantage in learning transformations between probability distributions. Hence, this section mainly discusses the implementation of probability distribution transformations by computing the OT map. Inspired by the AE-OT algorithm, a new solving algorithm is proposed, which is a reusable neural OT solver, its pipeline is shown in Fig. \ref{figure:OTnet}.

From Fig. \ref{figure:OTnet}, the key to our algorithm is learning Brenier's height representation using the network $H_{\Phi}$, where $\Phi$ is the network parameters. Specifically, we can randomly select some samples in the target domain and learn the height representation through $H_{\Phi}$. Then, we can use the learned Brenier's height representation to directly calculate the height vectors of the remaining samples. Even if new samples are added, the height vectors can be calculated straightforwardly using $H_{\Phi}$, without further training and optimization. Finally, Brenier's potential is obtained by substituting Eq. \eqref{eq:uh} whose gradient is the OT map. This greatly improves the computational efficiency and reusability of the OT map. Detailed numerical results are presented in section \ref{subsubsec3}.
\subsection{Learning Brenier's Height Representation} \label{sec:Height}
The core of calculating the OT map is to optimize the convex energy Eq. \eqref{eq:energy} to acquire the Brenier's height vector $\boldsymbol{h}$. AE-OT calculates the numerical solution of $\boldsymbol{h}$ under ad-hoc data and does not accept any change of data. That is to say, the height vector is not reusable, and the entire height vector $\boldsymbol{h}$ needs to be recalculated when new samples are added. To improve the reusability of the OT map and reduce the huge time cost of duplicate computing, we reformulate the support hyperplane $\pi_{\boldsymbol{h},i}(\boldsymbol{x})$ as follows,
\begin{equation}
\pi_{\boldsymbol{h},i}(\boldsymbol{x}) = \pi_{H(\boldsymbol{y}_i)}(\boldsymbol{x}) = \boldsymbol{x}^T \boldsymbol{y}_i + H(\boldsymbol{y}_i),
\label{eq:ourplane}
\end{equation}
\textcolor{blue}{where $H(\boldsymbol{y})$ is Brenier's height representation which is a function defined on the target domain $\boldsymbol{Y}$, and satisfies $\boldsymbol{h} = (h_1, h_2, ..., h_n)=(H(\boldsymbol{y}_1), H(\boldsymbol{y}_2), ..., H(\boldsymbol{y}_n))$. $n$ denotes the number of samples in the training dataset.}

\textcolor{blue}{Given $m$ new samples $\{\boldsymbol{y}_{n+1},\boldsymbol{y}_{n+2}...\boldsymbol{y}_{n+m}\}$ which are not included in the training dataset, and $r=\frac{n}{n+m}$ denotes the ratio between the training dataset $\{\boldsymbol{y}_{1},\boldsymbol{y}_{2}...\boldsymbol{y}_{n}\}$ and the whole dataset $\{\boldsymbol{y}_{1},\boldsymbol{y}_{2}...\boldsymbol{y}_{n+m}\}$. We directly input these $m$ new samples into the well-trained height representation network $H_{\Phi}$ to obtain the corresponding height. Then, Eq. \eqref{eq:ourplane} is substituted into Eq. \eqref{eq:uh} to obtain Brenier's potential $u_{\boldsymbol{h}}$:}
\begin{equation}
\textcolor{blue}{
u_{\boldsymbol{h}} = \max_{i=1}^{n+m}\{ \boldsymbol{x}^T \boldsymbol{y}_i  + H_{\Phi}(\boldsymbol{y_i})\}.}
\label{eq:ourbrenier}
\end{equation}

The gradient map $\nabla u_{\boldsymbol{h}}:\Omega\rightarrow\boldsymbol{Y}$ maps each cell $W_{i}(\boldsymbol{h})$ to a single
point $\boldsymbol{y}_{i}$:
\begin{equation}
\nabla u_{\boldsymbol{h}}:W_{i}(\boldsymbol{h})\rightarrow\boldsymbol{y}_i,\quad i=1,2,\ldots,n+m.
\end{equation}

Given the  empirical target measure $\nu$ which is the sum of the Dirac measures:
\begin{equation}
\nu=\sum_{i=1}^{n+m} \nu_{i} \delta\left(\boldsymbol{y}-\boldsymbol{y}_{i}\right), i=1,2, \ldots, n+m,
\end{equation}
there exists a discrete Brenier's potential in Eq.~\eqref{eq:ourbrenier} whose projected $\mu$ volume of each facet $W_{i}(\boldsymbol{h})$ is equal to the given target measure $\nu_i$. This was proved
by Alexandrov~\cite{alexandrov2005convex} in convex geometry.

\begin{algorithm}
\textcolor{blue}{
\caption{Brenier's Height Representation $H_{\Phi}$}\label{algorithm1}
\begin{algorithmic}[1]
\Require target sample points $\boldsymbol{Y}=\left\{\boldsymbol{y}_i\right\}_{i=1}^{n}$, the number of source samples $N$, termination criterion $\delta$, source distribution $\mathbb{P}_x$. 
\Ensure Brenier's Height Representation $H_{\Phi}$.
\State $\boldsymbol{h}=zeros(n)$. 
\While {$\| \nabla E(\boldsymbol{h})\|_2>\delta$}   
    \State $w(\boldsymbol{h})=zeros(n)$. 
    \State Sample $\boldsymbol{X}=\left\{\boldsymbol{x}_j\right\}_{j=1}^N\sim \mathbb{P}_x$ from source distribution.
    \State  $\boldsymbol{h} \leftarrow H_{\Phi}(\boldsymbol{Y})$.
    \State $\pi_{\boldsymbol{h},i}$ is calculated by Eq.(\ref{eq:ourplane}) for all $\boldsymbol{x}\in\boldsymbol{X}$, $\boldsymbol{{y}_i}\in\boldsymbol{Y}$.
    \State  Calculate the index of the maximum $\pi$, \textit{i.e.}, $index= k$ if $\max_{i=1}^{n}\{\pi_{\boldsymbol{h},i}\}=\pi_{\boldsymbol{h},k}$.
    \State \textbf{for} {$i=0;i < n$} \textbf{do} 
    \State ~~~~Count the number of occurrences of index $i$ and add it to $w_i(\boldsymbol{h})$.
    \State  ~~~~$i\leftarrow i+1$.
    \State \textbf{end}
    \State $w(\boldsymbol{h})\leftarrow w(\boldsymbol{h})/N$.
    \State  Calculate Eq.(\ref{eq:Ehgradient}) and  then use Adam \cite{adam2014} to optimize $H_{\Phi}$.
\EndWhile
\State return $H_{\Phi}$.
\end{algorithmic}
}
\end{algorithm}

Considering the universal approximation ability of the neural network, we use a neural network to fit height representation. Thanks to the continuous expression ability of neural networks, when new samples arrive in the target domain, \textcolor{blue}{we do not have to retrain model $H_{\Phi}$ by optimizing the energy function in Eq. (\ref{eq:energy}), nor do we need to recalculate the height vector of all samples. We just compute the height vector of the out-of-sample separately, which greatly improves the reusability of the algorithm. \textbf{Algorithm \ref{algorithm1}} summarizes the trained procedure of the network $H_{\Phi}$.}

\textcolor{blue}{When calculating the measure $w(\boldsymbol{h})$, the semi-discrete algorithm AE-OT~\cite{An2020AEOT} adopts the Monte Carlo method, which implicitly includes the process of discretizing continuous variables. This means that both our proposed \textbf{Algorithm \ref{algorithm1}} and AE-OT can be applied to semi-discrete as well as fully discrete cases. The difference between these two cases is whether the distribution $\mathbb{P}_x$ of the source data $\boldsymbol{X}$ is continuous or discrete. In the image generation and domain adaptation tasks, their source distributions are continuous Gaussian white noise and mixture Gaussian distribution, respectively, so they are semi-discrete OT problems. In the color transfer task, the $\mathbb{P}_x$ is discrete, so it is a fully discrete transport problem. In Section \ref{sec4}, we further elaborated and validated the applicability of the proposed algorithm in these tasks through experiments.}

\textcolor{blue}{In addition, the proposed algorithm accepts changes and additions to the data.} To be specific, when the new ad-hoc data are joined, the new samples and original samples fed to the \textbf{OT-Net} can directly acquire the whole height vector. Then, the support planes of $u_{\boldsymbol{h}}$ are obtained by Eq. (\ref{eq:ourplane}), and the potential is gained by taking the upper envelope of all supporting planes according to Eq. (\ref{eq:ourbrenier}). Finally, the OT map is given by the gradient map of potential. Unlike standard neural networks that are constrained to be continuous, the OT map from \textbf{OT-Net} can match any target distribution with many discontinuous supports and achieve sharp boundaries.

\subsection{Error Analysis} \label{sec:analysis}
\textcolor{blue}{In this section, we analyze the error bound of the presented algorithm. This error measures the distance between the height vector $\boldsymbol{h}$ obtained by training the height representation $H_{\Phi}$ with partial data $\boldsymbol{Y}_p =\{\boldsymbol{y}_{1},\boldsymbol{y}_{2}...\boldsymbol{y}_{n}\}$ and the exact height vector $\boldsymbol{h}^{\ast}$ obtained by training $H_{\Phi}$ with all data $\boldsymbol{Y}_a =\{\boldsymbol{y}_{1},\boldsymbol{y}_{2}...\boldsymbol{y}_{n},\boldsymbol{y}_{n+1}...\boldsymbol{y}_{n+m}\}$. Obviously, this error heavily depends on the distance between the newly added data $\{\boldsymbol{y}_{n+1},\boldsymbol{y}_{n+2}...\boldsymbol{y}_{n+m}\}$ and the manifold depicted by the original training data $\{\boldsymbol{y}_{1},\boldsymbol{y}_{2}...\boldsymbol{y}_{n}\}$. For further analysis, we assume $d(\boldsymbol{Y}_p ,\boldsymbol{Y}_a)\leqslant d_{max}$.}

To optimize the energy function $E(\boldsymbol{h})$ in \textbf{Algorithm \ref{algorithm1}}, the $\mu-$volume $w_{i}(\boldsymbol{h})$ of each cell $W_{i}(\boldsymbol{h})$ is the key step, which can be estimated using Monte Carlo method \cite{MC1949}. $N$ random samples is drew from distribution $\mathbb{P}_x$, then the $\mu-$volume of each cell is estimated as follows:
\begin{equation*}
    \tilde{w}_{i}(\boldsymbol{h}) = \mathbf{crad}\left( \{j\in \mathcal{J} | \boldsymbol{x}_j \in W_{i}(\boldsymbol{h})\} \right) /N, ~~~\mathcal{J}\in \{1,2,\cdots, N\}.
\end{equation*}
When $N$ is large enough, $\tilde{w}_{i}(\boldsymbol{h})$ converges to $w_{i}(\boldsymbol{h})$. Then, to minimize energy Eq. \eqref{eq:energy}, we learn Brenier's height representation $H(\boldsymbol{y})$ by parameterizing the neural network to obtain the optimal $\boldsymbol{h}$. Based on the properties of $\boldsymbol{h}$ mentioned in Ref. \cite{gu2016minkowski}, we propose the following proposition.

\begin{proposition}\label{theorem:convergent}
Let $\boldsymbol{Y}=\left\{\boldsymbol{y}_i\right\}_{i=1}^{n}$ be the feature set of the target samples, $\boldsymbol{X}=\left\{\boldsymbol{x}_j\right\}_{j=1}^N$ be source data sampled from source distribution $\mathbb{P}_x$. The height vector $\boldsymbol{h}= (H(\boldsymbol{y_1}), H(\boldsymbol{y_2}), ..., H(\boldsymbol{y}_n)) \in \mathbb{R}^n$ is obtained which minimizing the convex energy $E(\boldsymbol{h})$ of Eq.~\eqref{eq:energy} under the condition $\sum_i h_{i}=0$ through the \textbf{Algorithm \ref{algorithm1}}.  Then, it is generated sequence $\{\boldsymbol{h}^{(k)} \}$ which satisfy 
\begin{equation}
\begin{aligned}
      \boldsymbol{h}^{(k)} \rightarrow \boldsymbol{h}^{\ast}  ,~~ \mathbf{if}~~ \lim_{k\rightarrow \infty}\| \nabla E(\boldsymbol{h}^{(k)}) \|_2 = 0. \label{eq:convergent}   
\end{aligned}
\end{equation} 
Where $\boldsymbol{h}^{\ast}=(h^{\ast}_1, h^{\ast}_2, ..., h^{\ast}_n) \in \mathbb{R}^n $ is the exact solution of the energy function, it is existence and uniqueness.
\end{proposition}

According to the demonstration of Theorem 1.2 in Ref. \cite{gu2016minkowski}, it is clear that the above proposition holds.

Based on Proposition \ref{theorem:convergent}, $\| \boldsymbol{h} -\boldsymbol{h}^{\ast} \|_2 \leq \epsilon$ if $\| \nabla E(\boldsymbol{h})\|_2<\delta$, which is the termination criterion of the algorithm. Then we will further analyze the error bounds for obtaining out-of-sample height vectors using the well-trained $H_{\Phi}$ when new samples are joined. \textcolor{blue}{First, considering one new sample $\boldsymbol{y}_{n+1}$ is added. If this new sample is close to the manifold of the training data $d(\boldsymbol{y}_{n+1},\boldsymbol{Y}_p)\leqslant d_{max}$, then ${\exists}\ \tau>0$ makes $| H(\boldsymbol{y}_{n+1}) -  h^{\ast}_{n+1}| \leq \tau$ hold. Thereby the following inequality is yielded.}
\begin{equation}\label{eq:errorbound1}
    \| \boldsymbol{h} -  \boldsymbol{h}^\ast \|_2 \leq \sqrt{ (h_1- h_1^\ast)^2 +
    \cdots + (h_n- h_n^\ast)^2 + (H(\boldsymbol{y}_{n+1}) -  h^{\ast}_{n+1})^2 }
    = \sqrt{ \epsilon^2 + \tau^2}, 
\end{equation}
where $\epsilon$ and $\tau$ can be sufficiently small constants. $\boldsymbol{h}^\ast \in \mathbb{R}^{n+1}$ is the exact solution of energy Eq. \eqref{eq:energy}. $\boldsymbol{h} \in \mathbb{R}^{n+1}$ represents the height vector of $n+1$ samples, which is the result of combining  $H(\boldsymbol{y}_{n+1})$ into the optimal height vector in Proposition \ref{theorem:convergent}. Analogously, if $m$ samples are joined and nearly the manifold the target domain, the optimal $\hat{\boldsymbol{h}} \in \mathbb{R}^{m}$ of out-of-sample is obtained by the trained $H_{\Phi}$, its corresponding exact solution is $\hat{\boldsymbol{h}}^\ast \in \mathbb{R}^{m}$. We then give the following theorem.
\begin{theorem}
Let $ \boldsymbol{h}  := (\boldsymbol{h}, \hat{\boldsymbol{h}})\in \mathbb{R}^{n+m}$ and $ \boldsymbol{h} ^\ast := (\boldsymbol{h}^\ast, \hat{\boldsymbol{h}}^\ast) \in \mathbb{R}^{n+m}$ be the height vector of our method and the exact solution of Eq.~\eqref{eq:energy}, respectively. Then, the following inequality is valid
\begin{equation}\label{eq:errorbound2}
    \| \boldsymbol{h}  -  \boldsymbol{h} ^\ast \|_2 \leq \sqrt{ \epsilon^2 + m\tau^2}.
\end{equation}
\begin{proof}[proof]
According to the above derivation and Eq. \eqref{eq:errorbound1},  
 \begin{equation}
 \begin{split}
    \| \boldsymbol{h}  -  \boldsymbol{h}^\ast \|_2^2 & \leq  (h_1- h_1^\ast)^2 +
    \cdots + (h_n- h_n^\ast)^2 +  (\bar{h}_1- \bar{h}_1^\ast)^2 +
    \cdots + (\bar{h}_m- \bar{h}_m^\ast)^2 \\
    & \leq \epsilon^2 + m\tau^2.   
\end{split}
\end{equation}
Thus the Eq. \eqref{eq:errorbound2} holds.
\end{proof}
\end{theorem}
\textcolor{blue}{Combined with the above theoretical analysis, if the learned Brenier's height representation $H_{\Phi}$ is accurate enough and $m$ samples are adjacent to the manifold of the target domain, then the resulting height vector $\boldsymbol{h}$ will approximate the exact solution well within an upper bound given by Eq. \eqref{eq:errorbound2}. Although the upper bound depends on the regularity of $H_{\Phi}$ and the distance $d(\boldsymbol{Y}_p,\boldsymbol{Y}_a)$, We found experimentally that the relative error between $\boldsymbol{h}$ and  $\boldsymbol{h}^\ast$ is on the order of $10^{-3}$ even if $r$ is relatively small, e.g., $r = 0.7$.}

\section{Experiments}
\label{sec4}

To evaluate the performance of the \textbf{OT-Net}, we experiment with generative models, color transfer, and domain adaptation to show the effectiveness and efficiency of our algorithm. The detailed network architecture of the algorithm is showcased in Appendix \ref{secA1}. 

\subsection{Application to generative model} 
\begin{figure}[htb!]
  \centering
  \includegraphics[width=0.85\textwidth]{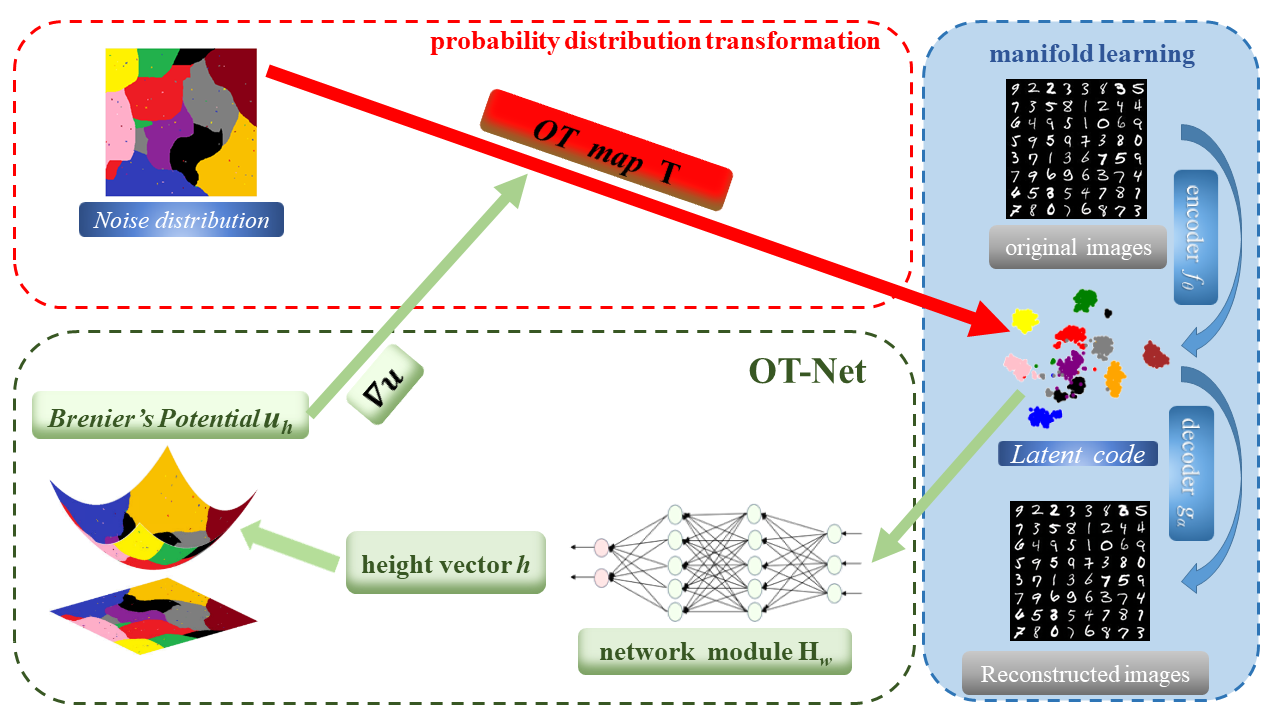}
  \caption{Framework for embedding \textbf{OT-Net} in generative models.  \textbf{manifold learning} (blue dotted box): $f_\phi$ and $g_\alpha$ represent the encoding and decoding map respectively, where $\phi$ and $\alpha$ are their network parameters. The illustrations of \textbf{OT-Net} (green dotted box) and \textbf{probability distributions transformation} (red dotted box) are detailed in the caption of Fig \ref{figure:OTnet}. }\label{figure:generativemodel}
\end{figure}
This section focuses on analyzing the performance of our algorithm when applied to the generative model. The diagram of our model is revealed in Fig \ref{figure:generativemodel}. Here, we use a network module to learn Brenier's height representation. Specifically, the latent codes are fed into the trained height representation to obtain its vector $\boldsymbol{h}$, which in turn induces the OT map. Finally, the generator of our model, which generates realistic images from random noise samples, is the composition $g_\alpha$ and $T$.

\textcolor{blue}{In the experiment, we first randomly select $n$ samples from the entire dataset at a given ratio $r$ to form the training dataset $\boldsymbol{Y}=\left\{\boldsymbol{y}_i\right\}_{i=1}^{n}$. Specifically, in Section \ref{subsubsec2}, we set $r=1$, and in Sections \ref{subsubsec1} and \ref{subsubsec3}, we selected several values for $r$ to demonstrate the reusability of the proposed method. Then, we obtain the height representation $H_{\Phi}$ by \textbf{Algorithm \ref{algorithm1}}. In the synthetic data experiment of Section \ref{subsubsec1}, the source distribution $\mathbb{P}_x$ in \textbf{Algorithm \ref{algorithm1}} is a normal distribution of the given mean and variance, while in Section \ref{subsubsec2} and \ref{subsubsec3}, $\mathbb{P}_x$ is the standard normal distribution.}

These experiments divide into three parts, in the first part, we conducted experiments on toy data to demonstrate that the proposed method can also avoid mode collapse/mixture. The second part evaluates the performance of our algorithm in the generative model. The final part confirms that our model can significantly enhance computational efficiency and reusability. Specifically, our algorithm is capable of training Brenier's height representation on a subset of the samples, and the height vectors of the remaining samples can be predicted by the trained model, ultimately obtaining the OT map.
These comparative experiments were conducted on four public datasets, \textit{i.e.}, \textbf{MNIST} \cite{MNIST}, \textcolor{blue}{\textbf{FASHION-MNIST}} \cite{Fashion}, \textbf{CIFAR-10} \cite{cifar10} and \textbf{CelebA} \cite{celebA}.

\subsubsection{Evaluation of Mode Collapse/Mixture in Synthetic Dataset}\label{subsubsec1}
The experiments focus on a synthetic dataset, as it has explicit distributions and known modes, allowing accurate measurement of mode collapse and the quality of generated samples. We selected the same synthetic dataset and metric indices as in \cite{lin2018pacgan}. Details are as follows.

\textbf{Dataset.} The \textbf{2D-ring} \cite{veegan} is a mixture of eight two-dimensional spherical Gaussians with means $(\cos ((2 \pi / 8) i), \sin ((2 \pi / 8) i))$ and variances $10^{-4}$ in each dimension for $i \in\{1, \ldots, 8\} .$ The \textbf{2D-grid} \cite{veegan} is a mixture of twenty-five two-dimensional spherical Gaussians with means $(-4+2 i,-4+2 j)$ and variances $0.0025$ in each dimension for $i, j \in\{0,1,2,3,4\}$.

\textbf{Metric.} To quantify the mode collapse behavior, we report three metrics: 
1)the \textbf{number of modes} counts the number of modes captured by samples generated from a generative model;
2)the \textbf{percentage of high-quality samples} is the ratio of such samples to the total number of synthetic data samples. If a sample falls within three standard deviations of the nearest mode, we consider it a high-quality sample;
3)the \textbf{reverse Kullback-Leibler divergence (reverse KL)} measures the balance between the induced distribution from generated samples and the induced distribution from the real samples.

\begin{table}[ht]

\caption{Experiments on \textbf{2D-ring} synthetic datasets under standard benchmark settings. $r$ indicates the proportion of the selected samples to the entire dataset. }\label{table:1}
\begin{tabular*}{\textwidth}{@{\extracolsep{\fill}}lccc@{\extracolsep{\fill}}}
\toprule%
& \multicolumn{3}{@{}c@{}}{2D-ring}  \\\cmidrule{2-4} %
& Modes(Max 8)  & high-quality samples & reverse KL   \\
\midrule
GAN\cite{Ugan2016} & $6.3 \pm 0.5$ & $98.2 \pm 0.2 \%$ & $0.45 \pm 0.09$   \\
ALI\cite{ALI2016} & $6.6 \pm 0.3$ & $97.6 \pm 0.4 \%$ & $0.36 \pm 0.04$    \\
MD\cite{MD2016}  & $4.3 \pm 0.8$ & $36.6 \pm 8.8 \%$ & $1.93 \pm 0.11$   \\
PacGAN2\cite{lin2018pacgan} & $7.9 \pm 0.1$ & $95.6 \pm 2.0 \%$ & $0.07 \pm 0.03$    \\
PacGAN3\cite{lin2018pacgan} & $7.8 \pm 0.1$ & $97.7 \pm 0.3 \%$ & $0.10 \pm 0.02$    \\
PacGAN4\cite{lin2018pacgan} & $7.8 \pm 0.1$ & $95.9 \pm 1.4 \%$ & $0.07 \pm 0.02$    \\
AE-OT\cite{An2020AEOT} & $\textbf{8.0} \pm \textbf{0.0}$ & $\textbf{99.6} \pm \textbf{0.3} \%$ &  $\textbf{0.004} \pm \textbf{0.001}$    \\
\midrule
Ours(r=0.9) & $8.0 \pm 0.0$ & $99.48 \pm 0.07 \%$ & $0.045 \pm 0.014$  \\
Ours(r=0.8) & $8.0 \pm 0.0$ & $98.72 \pm 0.23 \%$ & $0.239 \pm 0.073$    \\
Ours(r=0.7) & $8.0 \pm 0.0$ & $98.20 \pm 0.47 \%$ & $0.303 \pm 0.049$   \\
\bottomrule
\end{tabular*}

\end{table}

\begin{table}[ht]
\caption{Experiments on \textbf{2D-grid} synthetic datasets under standard benchmark settings. $r$ indicates the proportion of the selected samples to the entire dataset. }\label{table:2}
\begin{tabular*}{\textwidth}{@{\extracolsep{\fill}}lccc@{\extracolsep{\fill}}}
\toprule%
& \multicolumn{3}{@{}c@{}}{2D-grid} \\\cmidrule{2-4}%
& Modes(Max 25)  & high-quality samples& reverse KL \\
\midrule
GAN\cite{Ugan2016}  & $17.3 \pm 0.8$ & $94.8 \pm 0.7 \%$ & $0.70 \pm 0.07$ \\
ALI\cite{ALI2016}   & $24.1 \pm 0.4$ & $95.7 \pm 0.6 \%$ & $0.14 \pm 0.03$ \\
MD\cite{MD2016}  & $23.8 \pm 0.5$ & $79.9 \pm 3.2 \%$ & $0.18 \pm 0.03$ \\
PacGAN2\cite{lin2018pacgan}   & $23.8 \pm 0.7$ & $91.3 \pm 0.8 \%$ & $0.13 \pm 0.04$ \\
PacGAN3\cite{lin2018pacgan}  & $24.6 \pm 0.4$ & $94.2 \pm 0.4 \%$ & $0.06 \pm 0.02$ \\
PacGAN4\cite{lin2018pacgan}   & $24.8 \pm 0.2$ & $93.6 \pm 0.6 \%$ & $0.04 \pm 0.01$ \\
AE-OT\cite{An2020AEOT}  & $\textbf{25.0} \pm \textbf{0.0}$ & $\textbf{99.8} \pm \textbf{0.2} \%$ & $\textbf{0.007} \pm \textbf{0.002}$ \\
\midrule
Ours(r=0.9)  & $25.0 \pm 0.0$ & $99.62 \pm 0.02 \%$ & $0.0324 \pm 0.0004$ \\
Ours(r=0.8)   & $25.0 \pm 0.0$ & $99.37 \pm 0.02 \%$ & $0.0606 \pm 0.0002$ \\
Ours(r=0.7) & $25.0 \pm 0.0$ & $99.27 \pm 0.02 \%$ & $0.0932 \pm 0.0029$ \\
\bottomrule
\end{tabular*}

\end{table}
\begin{figure}[htb!]
\centering
\subfigure[real samples]{
\includegraphics[width=0.23\textwidth]{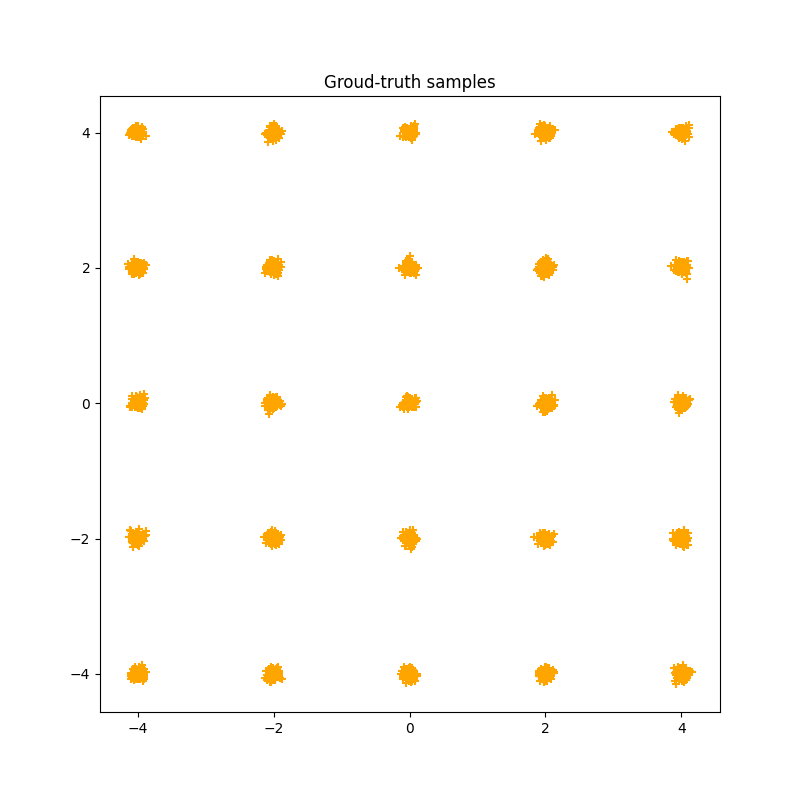}}
\subfigure[r=0.9]{
\includegraphics[width=0.23\textwidth]{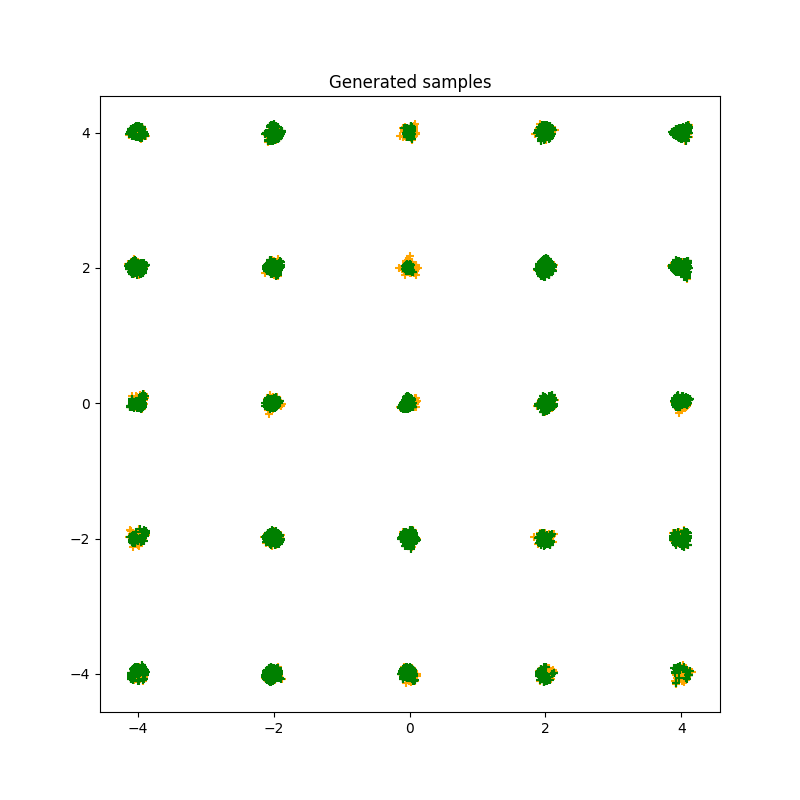}}
\subfigure[r=0.8]{
\includegraphics[width=0.23\textwidth]{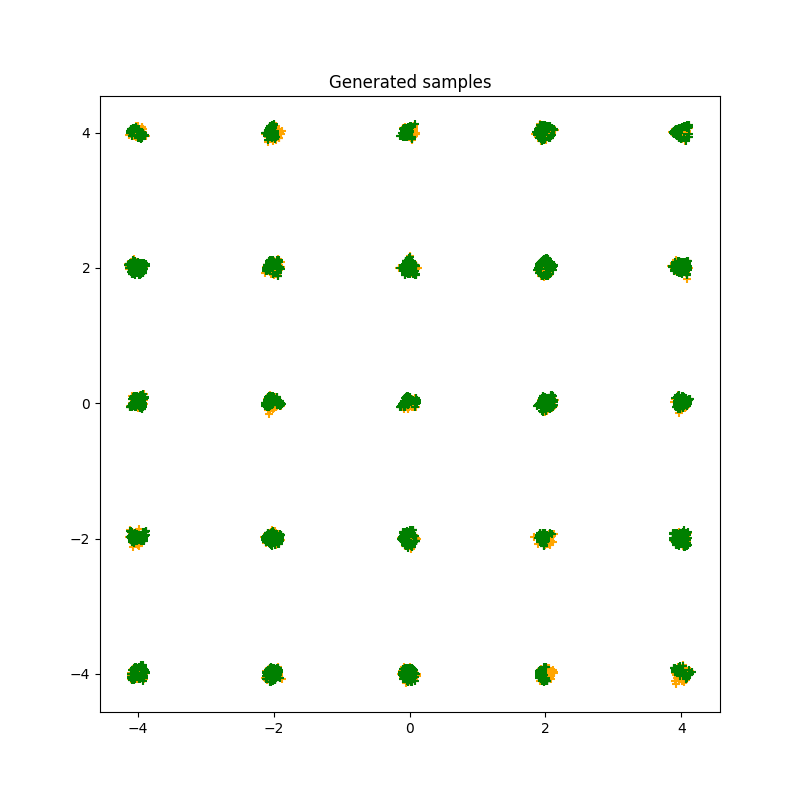}}
\subfigure[r=0.7]{
\includegraphics[width=0.23\textwidth]{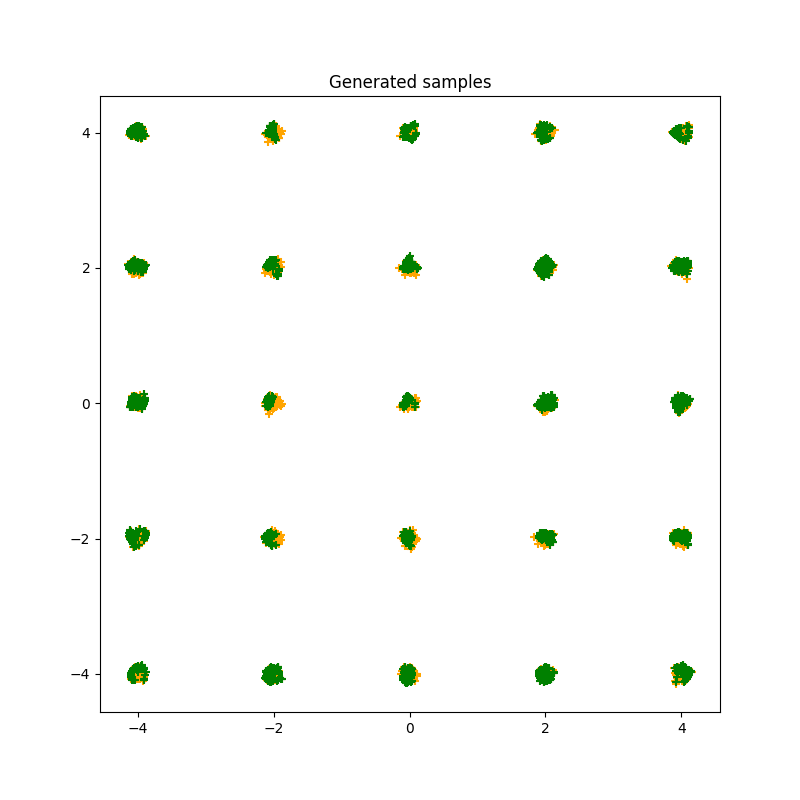}}
\caption{ Mode collapse/mixture comparison on the 2D grid dataset. Scatter plots are shown for 2D samples generated from (a) the true distribution of the 2D grid, and (b), (c), and (d) the generated distributions of the 2D grid using our model with different ratios of selected training samples. $r$ indicates the proportion of the training samples to the entire dataset. Orange marks represent real samples, and green marks represent generated samples. The results show that our model successfully captures all 25 modes.}\label{figure:videovisbd}
\end{figure}
Lin et al. \cite{lin2018pacgan} evaluated GAN, ALI, MD, and PacGAN on synthetic sets with the above three metrics. For the AE-OT and our model, there is no need to train an autoencoder because both the source domain and the target domain are two-dimensional. A two-dimensional extended OT map is straightforwardly computed.  But the way we calculate the OT map is different from AE-OT since the height vector is obtained through \textbf{OT-Net}. Our advantage is that we do not need to re-optimize the height representation. In other words, when adding new samples, \textbf{OT-Net} can directly provide the height vector of out-of-sample.

To verify the superiority of the proposed method, we randomly selected a portion of data from the entire dataset in ratios of 0.7, 0.8, and 0.9 to train the $H_{\Phi}$, and then used the well-trained $H_{\Phi}$ to calculate the OT map directly. The results are reported in Tab. \ref{table:1} $\sim$ Tab. \ref{table:2}, and results of previous methods are copied from Lin et al. \cite{lin2018pacgan} and An et al. \cite{An2020AEOT}. It can be seen that both the AE-OT and the proposed model outperform other models in these three evaluation metrics. \textcolor{blue}{From the Tab. \ref{table:1} $\sim$ Tab. \ref{table:2}, the percentage of high-quality samples and the reverse KL of our method are slightly below the AE-OT. This is due to our algorithm providing an approximate solution for OT map compared to AE-OT. Yet, our method still performs well in avoiding mode collapse.} The results indicate that our method can capture all models and generate higher-quality samples, even when the number of features is limited. Furthermore, the visual results for 2D-grid data are displayed in Fig. \ref{figure:videovisbd}, which indicates that our approach can capture all modes without mode mixture when selecting different samples of different proportions. Based on these results, it can be concluded that the proposed model effectively addresses mode collapse/mixture issues. What's more, this shows the height representation learned with part of the samples can predict the out-of-sample height.

\subsubsection{Quality Evaluation of Generative Image}\label{subsubsec2}
To evaluate the performance of our algorithm used for image generation, we perform quantitative and qualitative experiments.
Fréchet Inception Distance (FID) was proposed by Heusel et al. \cite{heusel2017gansfid} to quantify the quality of generated samples.


\begin{table*}[h]
\centering
\begin{minipage}{\textwidth}
\caption{Quantitative comparison with FID $\downarrow$. The best result is shown in bold. MM: manifold matching; NS: non-saturating; LSGAN: least-squares GAN; BEGAN: boundary equilibrium GAN.}\label{table:fid}
\scalebox{0.6}{
\begin{tabular}{lccccccccc} 
\toprule%
Dataset  & MM GAN \cite{mmns2018} & NS GAN\cite{mmns2018} & LSGAN\cite{ls2017}& WGAN\cite{wgan2017}& BEGAN\cite{be2017}& VAE\cite{VAE2013}& GLO\cite{GLO2017}& AE-OT\cite{An2020AEOT} & Ours  \\
\midrule
\textcolor{blue}{FASHION-MNIST} & 29.6 &26.5 &30.7 &21.5 &22.9 &58.7& 57.7 &10.98 & \textbf{9.85}  \\
Cifar-10  & 72.7 &58.5 &87.1 &55.2 &71.4 &155.7 &65.4 & 25.98  & \textbf{24.09} \\
CelebA & 65.6 &55.0 &53.9 &41.3 &38.9 &85.7 &52.4  & 38.68   & \textbf{34.83}\\
\bottomrule 
\end{tabular}
}
\end{minipage}
\end{table*}

Depending on its calculation formula, the FID scores are reported in Tab. \ref{table:fid}, and statistics of various GANs come from Ref. \cite{An2020AEOT,hoshen2018non,lucic2018gans}. The results of the AE-OT and our model are obtained under the same Encoder-Decoder architecture. \textcolor{blue}{We have re-optimized the AE-OT model and achieved better results than those presented in table 2 of AE-OT~\cite{An2020AEOT}. 
Through fine-tuning the model and adjusting hyperparameters, we achieved a lower FID score compared to AE-OT. Note that the key idea of this paper is to introduce a reusable method for OT computation. The integration of this method with generation, color transfer, and domain adaption aims to exemplify the scalability and efficacy of the proposed algorithm.}

\begin{figure}[htb!]
\centering
\subfigure[VGAN~\cite{zhai2016generative}]{
\includegraphics[width=0.18\textwidth]{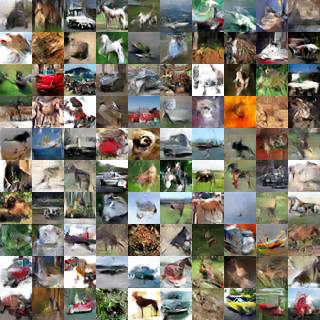}}
\subfigure[BCGAN~\cite{abbasnejad2017bayesian}]{
\includegraphics[width=0.18\textwidth]{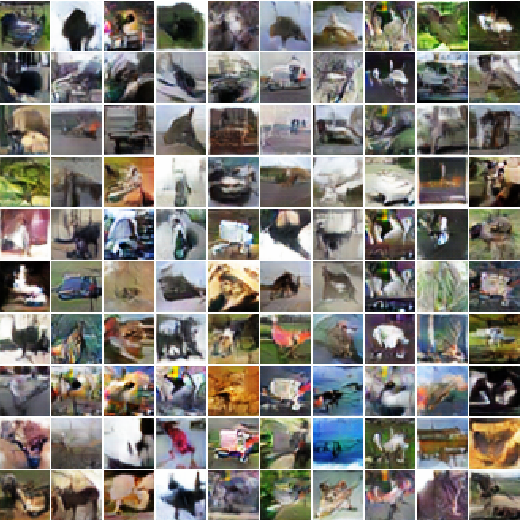}}
\subfigure[WGAN-GP~\cite{gulrajani2017improvedgan}]{
\includegraphics[width=0.18\textwidth]{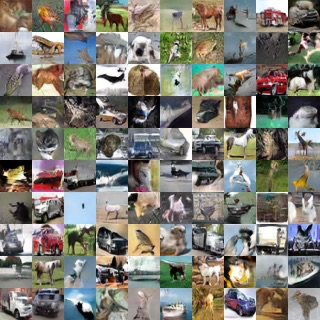}}
\subfigure[AE-OT~\cite{An2020AEOT}]{
\includegraphics[width=0.18\textwidth]{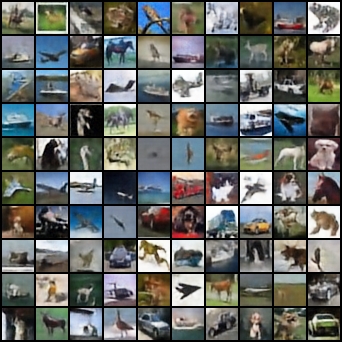}}
\subfigure[OT-Net]{
\includegraphics[width=0.18\textwidth]{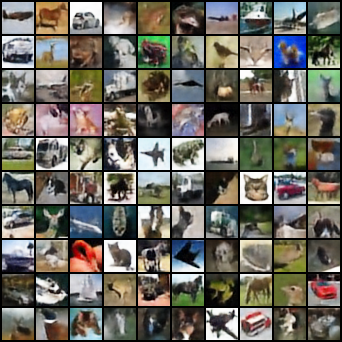}}
\caption{The visual comparison on Cifar10 dataset.}\label{figure:cifar10}
\end{figure}

\begin{figure}[htb!]
\centering
\subfigure[$\alpha$-GAN~\cite{rosca2017variational}]{
\includegraphics[width=0.18\textwidth]{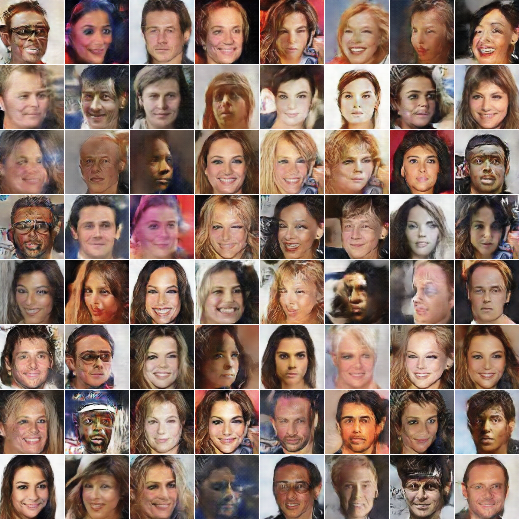}}
\subfigure[VEEGAN~\cite{veegan}]{
\includegraphics[width=0.18\textwidth]{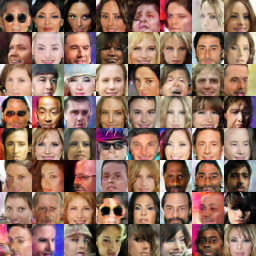}}
\subfigure[WGAN-QC~\cite{liu2019wganqc}]{
\includegraphics[width=0.18\textwidth]{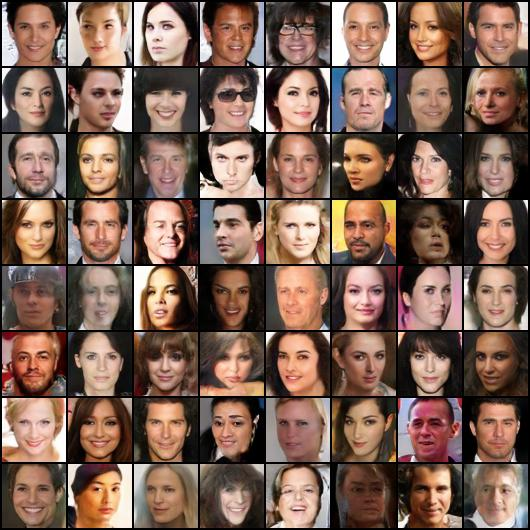}}
\subfigure[AE-OT~\cite{An2020AEOT}]{
\includegraphics[width=0.18\textwidth]{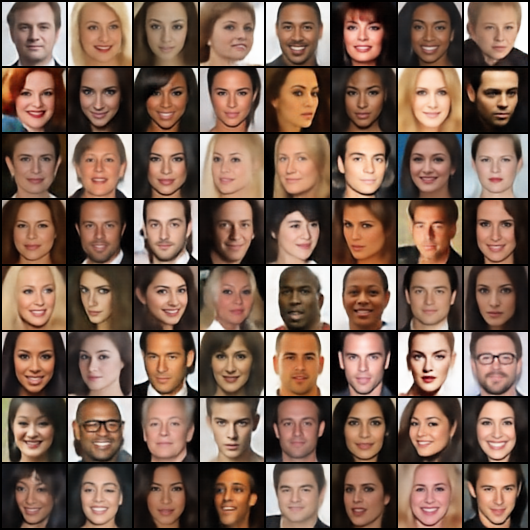}}
\subfigure[OT-Net]{
\includegraphics[width=0.18\textwidth]{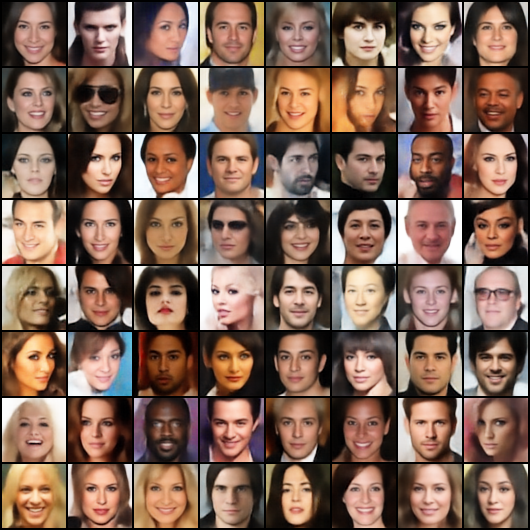}}
\caption{The visual comparison on CelebA dataset.}\label{figure:celeba}
\end{figure}

\begin{figure}[htb!] 
\begin{center}
\subfigure[OT-ICNN\cite{Makkuva2020}]{
\includegraphics[width=0.24\textwidth]{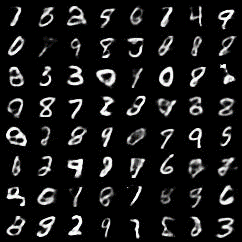}}
\subfigure[OTM\cite{Rout2021}]{
\includegraphics[width=0.24\textwidth]{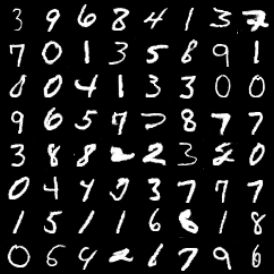}}
\subfigure[OT-Net]{
\includegraphics[width=0.24\textwidth]{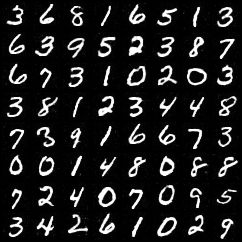}}
\end{center}
\caption{The visual comparison of neural-network-based OT algorithms on MNIST.}\label{figure:NNOT}
\end{figure}

Furthermore, we provide \textbf{qualitative comparisons} between our method and various GAN models.
Fig. \ref{figure:cifar10} $\sim$ Fig. \ref{figure:celeba} depict visual comparisons between images generated by our model and the reported results of other models. From the visual results, the generative images produced by our proposed method are of better quality compared to other models. Moreover, the generated face images are not blurred or mixed.

\begin{figure}[htb!] 
\begin{center}
\subfigure{
\includegraphics[width=0.75\textwidth]{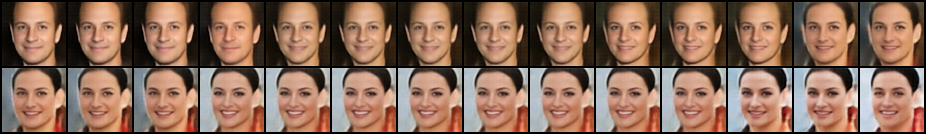}}
\end{center}
\caption{ Our method for latent space interpolation on CelebA (64 × 64).}\label{figure:interpolation}
\end{figure}

In addition to the aforementioned results, our algorithm is compared with other \textcolor{blue}{neural network-based OT algorithms.} Namely, Makkuva et al. \cite{Makkuva2020} proposes a new framework called OT-ICNN, which utilizes Input Convex Neural Networks to estimate the gradient of a convex function as an OT map; Rout et al. \cite{Rout2021} present an end-to-end algorithm (OTM) for fitting OT map with quadratic cost. Results are shown in Fig. \ref{figure:NNOT}, where we can observe that the generated image results of OT-ICNN and OTM exhibit mode mixing, whereas our algorithm does not. Moreover, we display the latent space interpolation between the generated samples in  Fig. \ref{figure:interpolation}, which also shows our model can avoid mode mixture.


\subsubsection{Verifying Reusability of OT-Net in Image Generation}\label{subsubsec3}

\begin{table}[htb!]
\caption{A comparison was made between the time cost of AE-OT and OT-Net for computing OT maps using the same encoder-decoder architecture. The training sample ratio indicates the proportion of selected samples. The results demonstrate that this method significantly improves computational efficiency. }\label{table:time}
\begin{tabular*}{\textwidth}{@{\extracolsep{\fill}}ccccccc@{\extracolsep{\fill}}}
\toprule%
Training  & \multicolumn{2}{c}{FASHION-MNIST} & \multicolumn{2}{c}{Cifar} & \multicolumn{2}{c}{CelebA} \\ \cmidrule{2-3}\cmidrule{4-5}\cmidrule{6-7}
feature ratio & AE-OT (s) & Ours (s) & AE-OT (s) & Ours (s) & AE-OT (s) & Ours (s)\\
\midrule
 1.0 &134.83  & \textbf{37.22} &222.82  & \textbf{21.97}	& 417.73  & \textbf{57.31}\\
  0.95 & -- & 24.56 & -- & 28.11 & -- &32.64\\
  0.90 & -- & 26.62 & -- & 19.81 & --&36.16\\
  0.85 & -- & 29.30 & -- & 17.27 & --&37.33\\
  0.80 & -- & 19.61 & -- & 28.49 & --&36.31 \\
  0.75 & -- & 23.42 & -- & 24.42 & --&32.35 \\
  0.70 & -- & 16.35 & -- & 22.90 & --&44.15 \\
\bottomrule
\end{tabular*}
\end{table}

In this section, we will highlight the strengths of our algorithm, \textit{i.e.}, we can train the Brenier's height representation with a portion of data, and then use the trained model to predict the height vectors of the remaining samples, and finally obtain the OT map directly. Compared to the AE-OT algorithm, the \textbf{OT-Net} is reusable. To verify the reusability and effectiveness of the proposed algorithm, we apply it to the generative model in our experiment.

\begin{figure}
    \centering
    \includegraphics[width=1.0\textwidth]{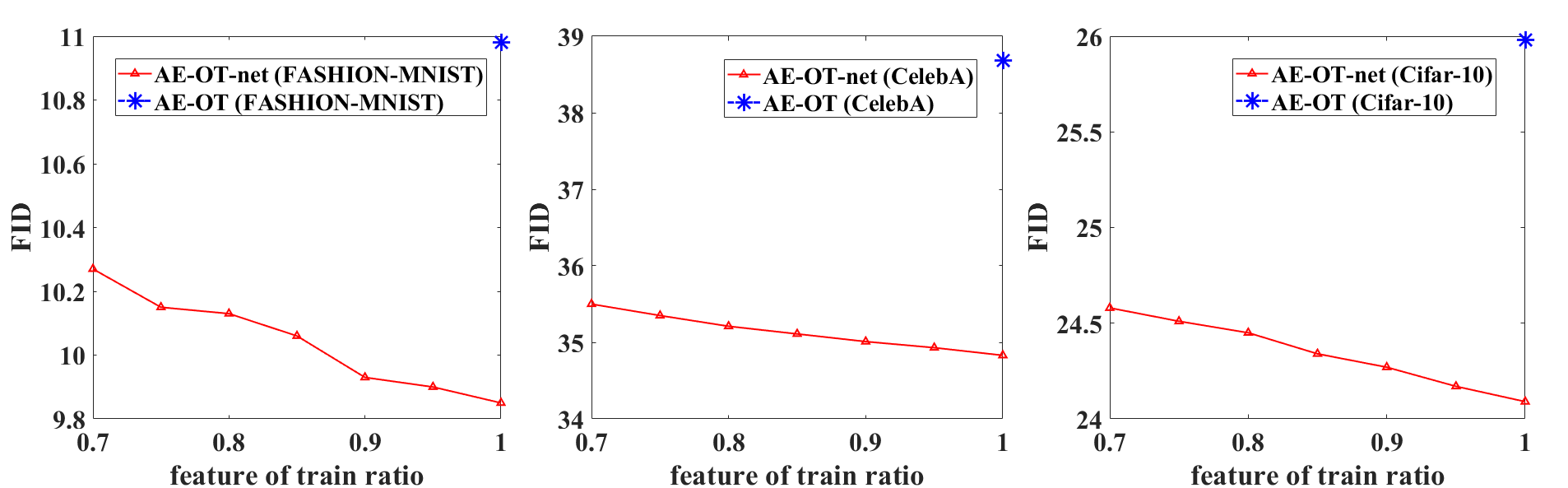}
    \caption{The curves on the left, middle, and right display the visual of FID scores with different feature ratios selected on the FASHION-MNIST, CelebA, and Cifar-10, respectively.}\label{figure:fashionceleba}
\end{figure}

\begin{figure}[htb!]
\begin{center}
\subfigure{
\includegraphics[width=0.23\textwidth]{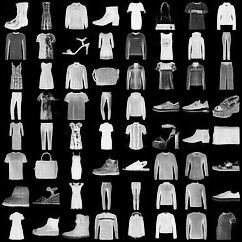}}
\subfigure{
\includegraphics[width=0.23\textwidth]{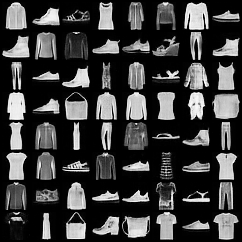}}
\subfigure{
\includegraphics[width=0.23\textwidth]{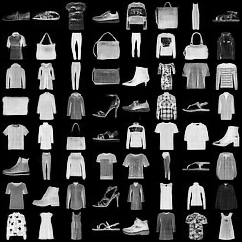}}
\subfigure{
\includegraphics[width=0.23\textwidth]{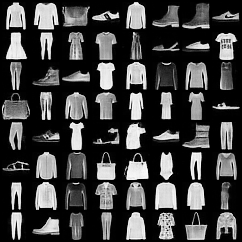}}
\subfigure{
\includegraphics[width=0.23\textwidth]{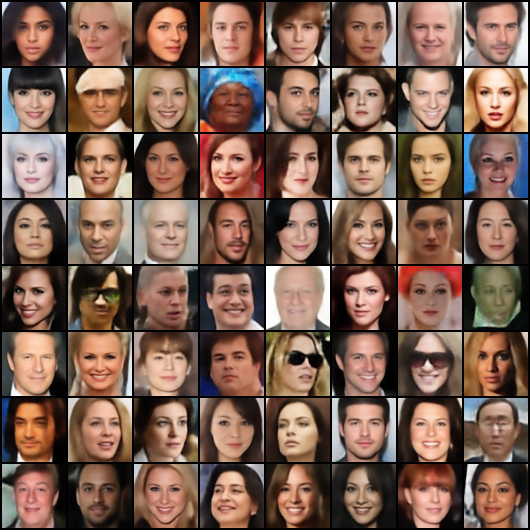}}
\subfigure{
\includegraphics[width=0.23\textwidth]{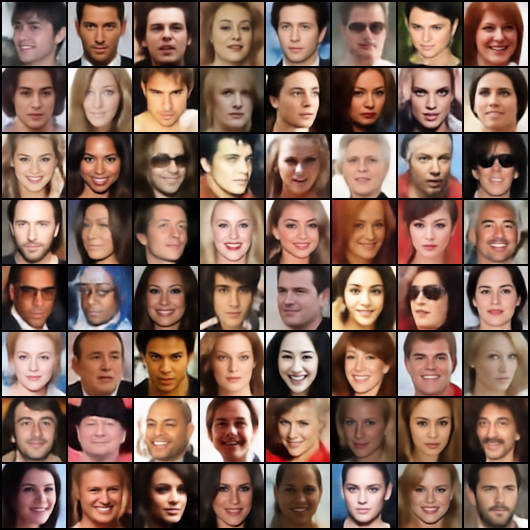}}
\subfigure{
\includegraphics[width=0.23\textwidth]{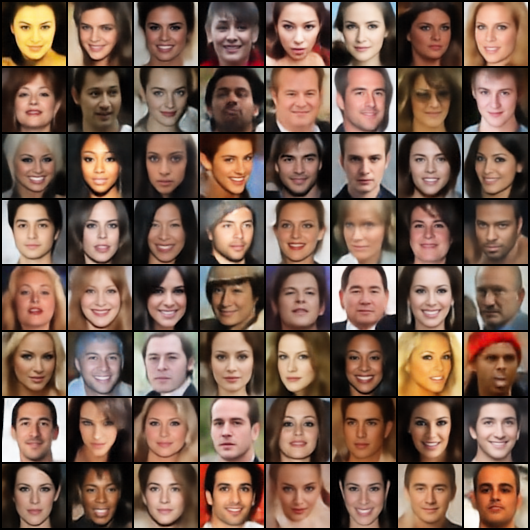}}
\subfigure{
\includegraphics[width=0.23\textwidth]{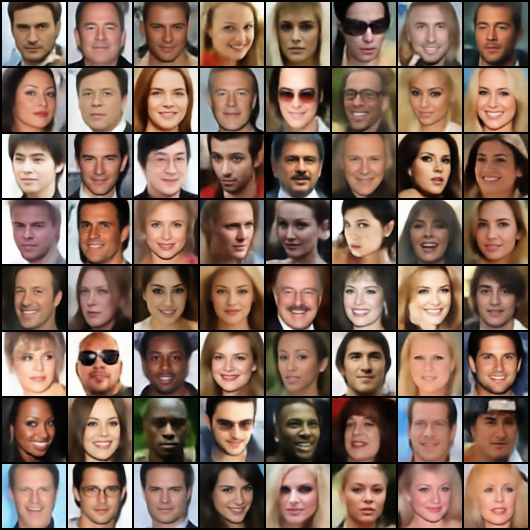}}
\end{center}
\caption{The generated images by our model were visually compared under different training sample ratios $r$. The ratios of selected samples on FASHION-MNIST and CelebA are 0.7, 0.8, 0.9, and 1.0, from left to right.}\label{figure:aeotnetfashion}
\end{figure}

Firstly, based on the results revealed in Fig. \ref{figure:fashionceleba}, although the FID score gradually increases as the number of selected features decreases, the image quality is still superior to those obtained by AE-OT. As presented in Tab. \ref{table:fid}, the network enables the computation of the whole height vector using a randomly selected subset of features for training. In case of new feature additions, the trained network can directly calculate the height vector to obtain the OT map. The visual effects of the newly generated images are shown in Fig. \ref{figure:aeotnetfashion}. Where $0.7 \sim 1.0$ indicates the ratio of randomly selected samples to the entire dataset.
\textcolor{blue}{OT-Net can train the Brenier's height representation with a portion of the dataset to train the model, its ultimate goal is to calculate the OT map for the entire dataset. This improves the efficiency and reusability of the algorithm. In AE-OT, if a portion of the data is selected for calculating the OT map, then only the OT map for that specific subset of data can be obtained. It cannot be directly extrapolated to the OT map for the entire dataset. Instead, the overall OT map needs to be recalculated. Considering these, we did not use AE-OT to calculate partial OT maps when $r$ is not equal to 1.}
 
Secondly, in Tab. \ref{table:time}, the time taken to train the OT map is reported for different training sample ratios $r$. The time reduction is not exact as the $r$ decreases, due to the discrepancy in the number of iteration steps required for the algorithm to converge under the same error threshold. When all the features are fed into the network for training, the proposed approach only takes 37.22s, 21.97s, and 57.31s to solve the OT map on the FASHION-MNIST, Cifar-10, and CelebA datasets, respectively. Compared with the AE-OT model, the time is greatly reduced. To be exact, once new features are added, we can readily calculate the height vector using the trained height representation. When we select $80\%$ of the data samples, calculating the height vector of the remaining samples takes only 0.028s on the MNIST-fashion, similarly, 0.078s on Cifar-10 and 0.513s on CelebA.
However, when encountering the above situation, the AE-OT model needs to be retrained, which inevitably results in increased computational costs.
The time taken for training OT is reported in Tab. \ref{table:time}, which is 134.83s, 222.82s, and 417.73s for the FASHION-MNIST, Cifar-10, and CelebA datasets, respectively. In summary, the \textbf{OT-Net} not only significantly reduces the training time of OT, but also is capable of predicting the height vectors of out-of-sample, thereby directly getting the OT map.

\begin{figure}[htb!]
\centering
\subfigure{
\includegraphics[width=0.48\textwidth]{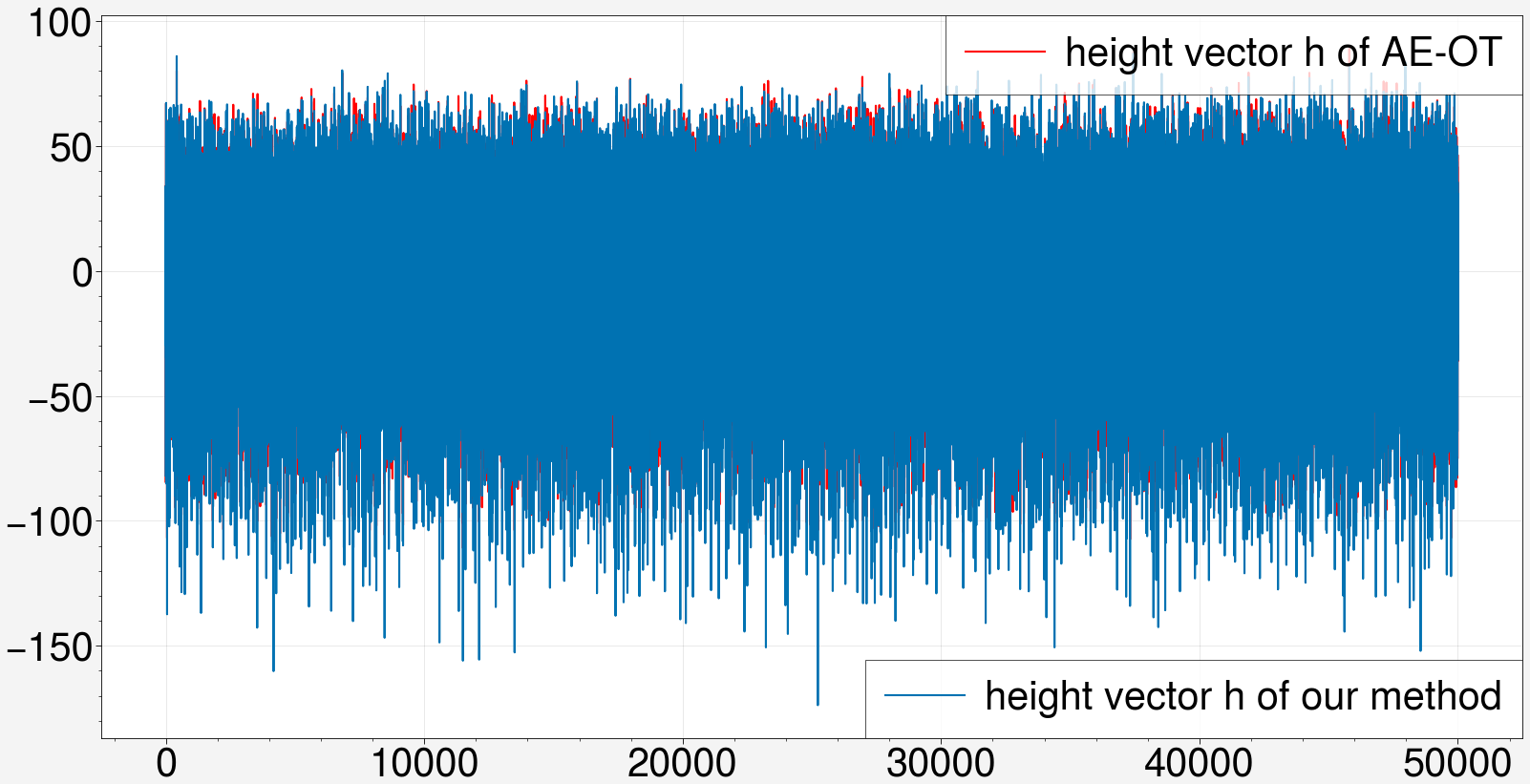}} 
\subfigure{
\includegraphics[width=0.48\textwidth]{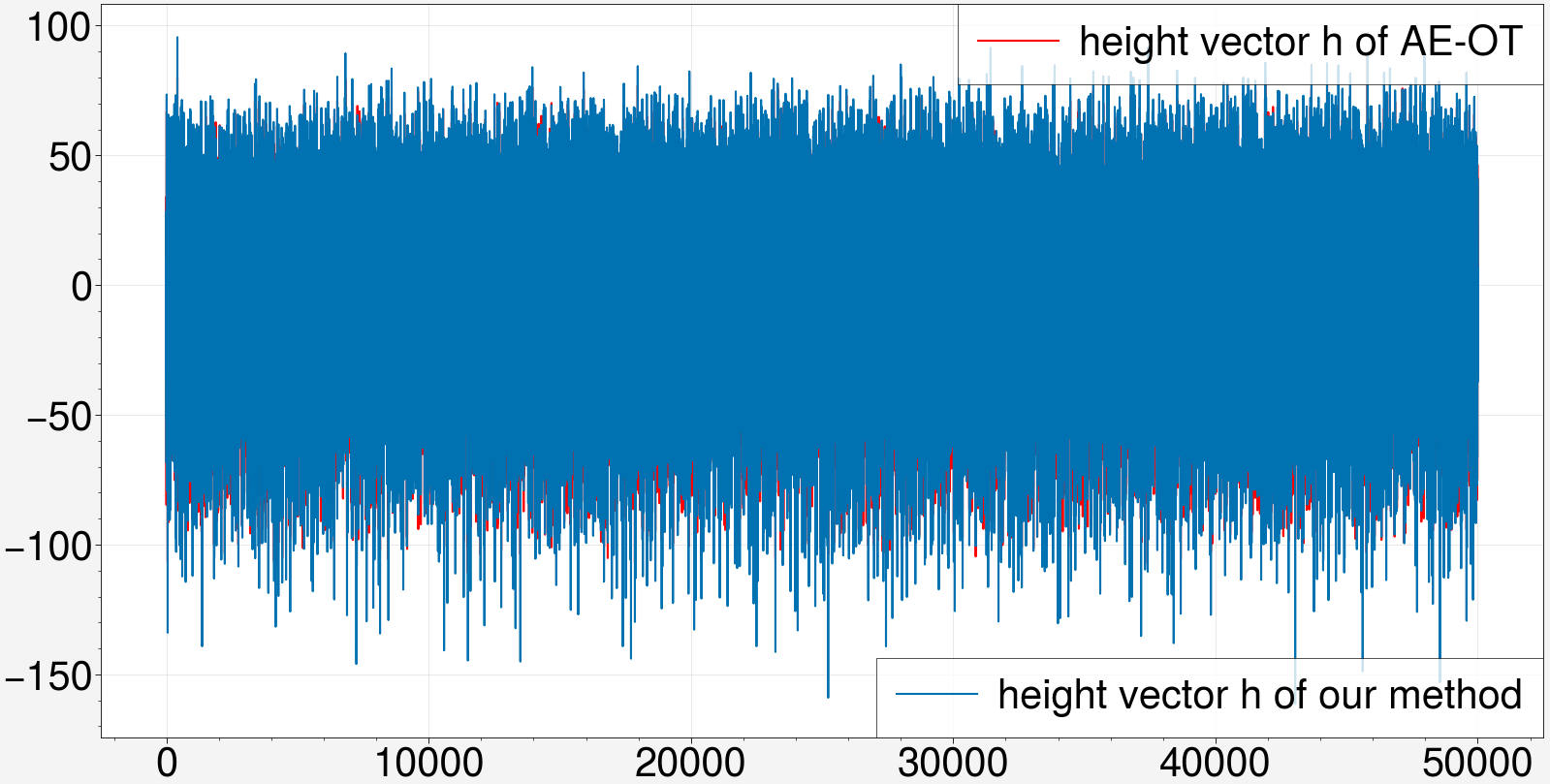}}
\subfigure{
\includegraphics[width=0.48\textwidth]{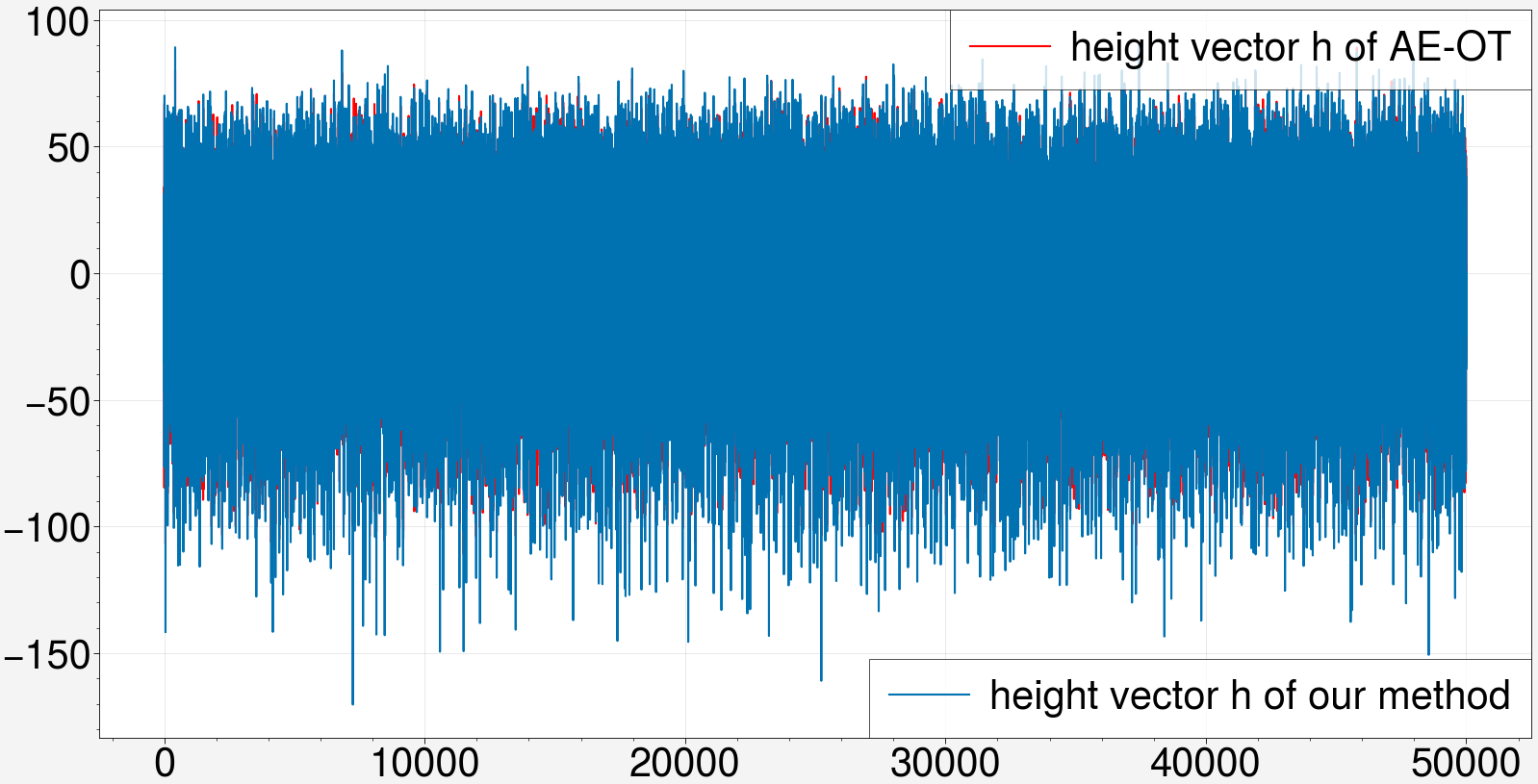}}
\subfigure{
\includegraphics[width=0.48\textwidth]{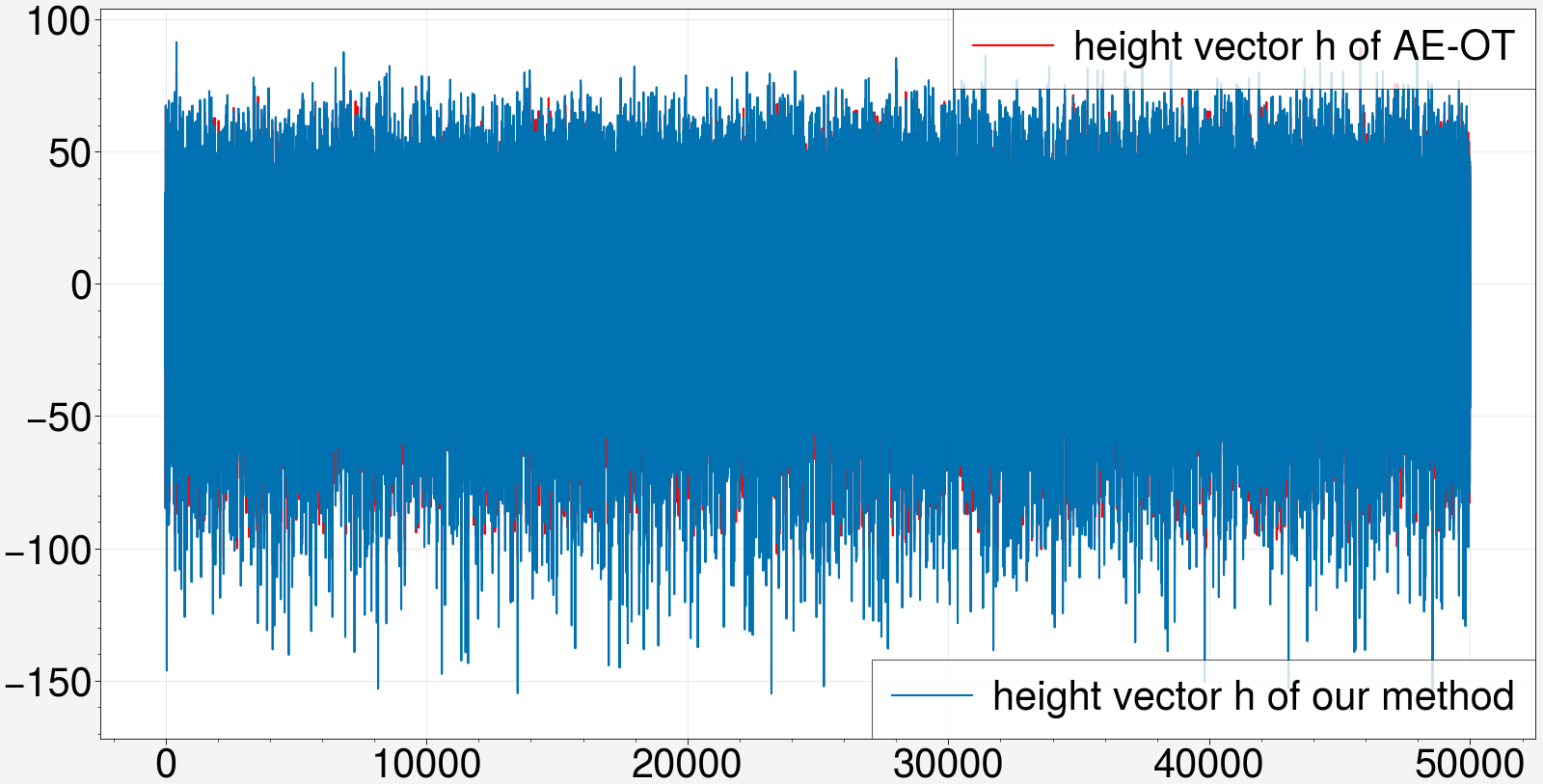}}
    \caption{Visual comparisons of height vectors $\boldsymbol{h}$ obtained by AE-OT and our OT-Net. In the subfigures of Top left, top right, bottom left and bottom right, our OT-Net selects 70\%, 80\%, 85\%, and 95\% of the entire CIFAR-10 dataset, respectively, to train the model to obtain the Brenier's height representation $H_{\Phi}$. Then given the entire dataset $\boldsymbol{Y}$, the height vector $\boldsymbol{h}$ is directly output by $H_{\Phi}$.}\label{fig:ratio}
\end{figure}

\textcolor{blue}{Moreover, to further certify the stability of our OT-Net, we compared the height vectors $\boldsymbol{h}$ computed by our OT-Net and AE-OT \cite{An2020AEOT}, which are reported in Fig. \ref{fig:ratio}. The proposed OT-Net first randomly selects a portion of data from the entire CIFAR-10 dataset to learn the brenier's height representation, and then directly computes the entire height vector $\boldsymbol{h}$. The results of AE-OT were optimized on the entire dataset and can be taken as a ground truth. From Fig. \ref{fig:ratio}, the height vectors computed by our algorithm closely approximate that computed by AE-OT. This indicates that the estimation of the OT map remains accurate, even when adding out-of-sample points.}

Finally, the experimental results show that the algorithm can not only directly predict out-of-sample height vectors accurately using the trained height representation, but also can significantly improve the computational efficiency of the algorithm.

\subsection{ Application to color transfer} 
\begin{figure}[htb!]
\centering
\subfigure{
\includegraphics[width=0.7\textwidth]{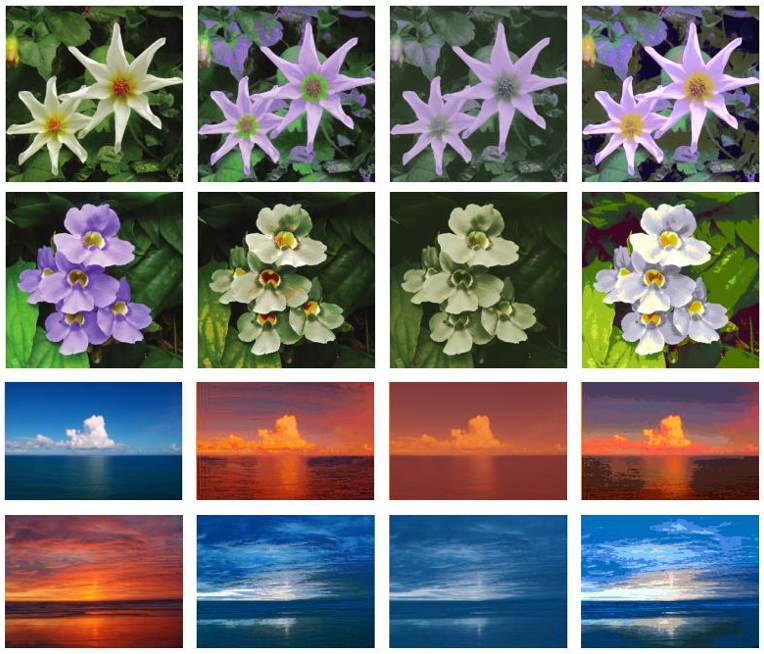}} 
\caption{Comparison of the visual results of color transfer. The first and third columns are original images, and the second and fourth columns are the results of the OT Network Simplex solver \cite{ns2011}, Sinkhorn algorithm \cite{cuturi2013sinkhorn}, and \textbf{OT-Net}, respectively.}\label{figure:colortransfer}
\end{figure}

This section will present comparative experiments on color transfer between our algorithm and other classic OT algorithms. The color transfer problem is to modify an input image $X_0$ so that its colors match the colors of another input image $Y_0$. 

\textcolor{blue}{For our algorithm implementation, we first randomly select $n$ pixels from target style image at a given ratio $r$ to form the training dataset $\boldsymbol{Y}=\left\{\boldsymbol{y}_i\right\}_{i=1}^{n}$. Then, we obtain the height representation $H_{\Phi}$ by \textbf{Algorithm \ref{algorithm1}}. Here, the source distribution $\mathbb{P}_x$ in \textbf{Algorithm \ref{algorithm1}} is a uniform distribution, so that the probability of each pixel being selected is equal. The image size is $512 \times 512$, $n=500$, $N=500$.}

For a fair comparison, we use the solver provided in the POT\footnote{https://pythonot.github.io/\#} \cite{pot2021}. Specifically, OT Network Simplex solver \cite{ns2011} for calculating Earth Movers Distance(EMD), and Entropic regularization OT solver with Sinkhorn-Knopp Algorithm \cite{cuturi2013sinkhorn}. 
Furthermore, we experiment on the same hardware platform and environments, the regularized coefficient of the Sinkhorn algorithm is 0.1, and the learning rate of our algorithm is 0.05. We selected two color styles of flowers and oceans for the color transfer task. The visual results are displayed in Fig. \ref{figure:colortransfer}, which reveals our algorithm generates sharper textures with more realistic and rich colors, especially in the background areas.

 Specifically, the first two rows of Fig. \ref{figure:colortransfer} show the results of color transfer between the white and purple flowers, where the result of \textbf{OT-Net} in the last column reveals that the purple nicely transfers to the white flowers. The last two rows represent the color transfer from the ocean blue sky with white clouds to the red sunset, where the last column shows that \textbf{OT-Net} is capable of transforming the colors of both scenes very effectively. The second and third columns display the results of the OT Network Simplex solver and Sinkhorn algorithms, respectively. We found that the image appears dull and the colors are not bright enough after color transfer. \textcolor{blue}{In addition, we compared the running times of our method with several other standard solvers and presented the results in Tab. \ref{table:time2}. This shows our algorithms are efficient and consume minimal time.  }

\begin{table}[htb!]
\centering
\caption{Comparison of running times of three OT solvers applied to color transfer. }\label{table:time2}
\begin{tabular}{l|c|c|c} 
\toprule 
  & Network Simplex solver \cite{bonneel2011displacement} & Sinkhorn algorithm \cite{cuturi2013sinkhorn} & OT-Net \\  
\midrule
time  &    1.777s          &  5.587s               &  \textbf{1.102s}   \\
\bottomrule
\end{tabular}
\end{table}



\subsection{Application to domain adaptation} 
\begin{figure}[htb!]
\centering
\includegraphics[width=0.9\textwidth]{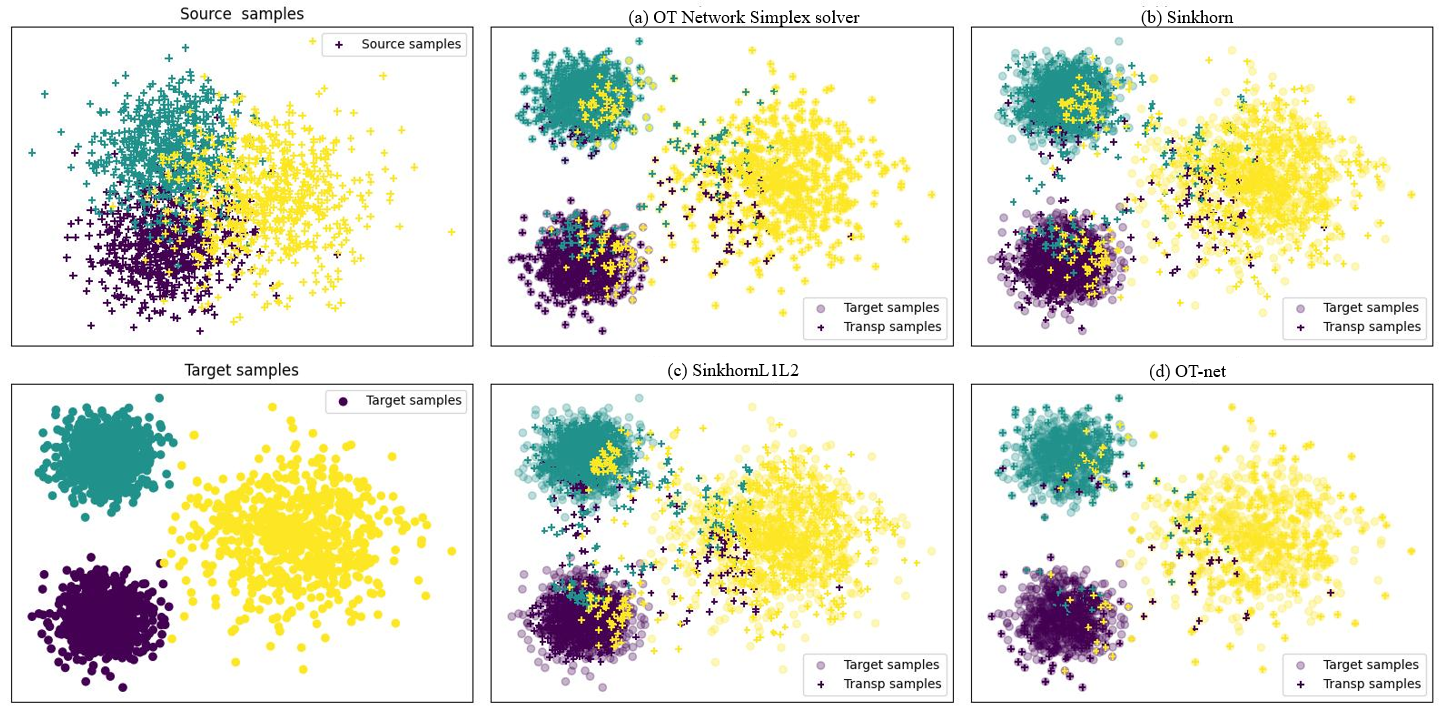}
\caption{Comparison of the visual results of domain adaption. The domain adaptation accuracies are (a): OT Network Simplex solver \cite{ns2011}, $83.55\%$, (b): Sinkhorn algorithm \cite{cuturi2013sinkhorn}, $79.49\%$, (c): SinkhornL1L2 \cite{OTDA2016}, $81.20\%$, (d): our \textbf{OT-Net}, $\mathbf{84.45\%}$.}\label{figure:OTDA}
\end{figure}

This part mainly introduces the applications of our algorithm in domain adaptation. \textcolor{blue}{For our algorithm implementation, we first randomly select $n$ samples from target domain at a given ratio $r$ to form the training dataset $\boldsymbol{Y}=\left\{\boldsymbol{y}_i\right\}_{i=1}^{n}$. Then, we obtain the height representation $H_{\Phi}$ by \textbf{Algorithm \ref{algorithm1}}. Here, the source distribution $\mathbb{P}_x$ in \textbf{Algorithm \ref{algorithm1}} is a mixture Gaussian distribution. The total number of samples in the target domain is $4000$, and $n=N=4000r$.}

In the 2D toy dataset setting, we randomly generate 4000 sample points and divide them into 3 classes as source domain (target domain) which are generated by different Gaussian mixture models. Three classical algorithms are selected as baselines. The first one is proposed by Flamary et al. \cite{OTDA2016} which adds a class-based term to the regularized OT for domain adaptation, shorthand for SinkhornL1L2, where L1 and L2 represent the regularization term of OT and class, respectively. The coefficients of OT and class-based regularization terms are 1.0 and 0.1. The remaining two are the OT Network Simplex solver \cite{ns2011} for computing EMD, and the Sinkhorn-Knopp algorithm \cite{cuturi2013sinkhorn} that the regularization coefficient is 0.1. We evaluate the performance of all those algorithms by calculating the percentage of correctly classified instances across all categories. From the results shown in Fig. \ref{figure:OTDA}, the proposed algorithm obtains the highest accuracy $\mathbf{84.45\%}$, this shows that our algorithm is effective. 

\begin{table*}[htb!]
\centering
\caption{Classification accuracy of domain adaptation. The first row represents the ratio of selected samples used for training. Part-data-Acc of the second row denotes the classification accuracy of the proposed method on these selected samples. All-data-Acc of the third row denotes the classification accuracy of the proposed method on the entire dataset. The two accuracies in the same column are obtained using the same trained $H_{\Phi}$. }\label{table:acc}
\begin{tabular}{lcccccc} 
\toprule 
Data-Ratio  & $70\%$ & $75\%$ & $80\%$ & $85\%$  & $90\%$ & $95\%$  \\  
\midrule
Part-data-Acc  & $83.46\%$  &$83.83\%$   &$83.87\%$  &$84.15\%$  &$84.25\%$  &$84.61\%$ \\
All-data-Acc   & $83.18\%$  &$83.55\%$   &$83.65\%$  &$84.11\%$  &$84.18\%$  &$84.33\%$  \\
\bottomrule
\end{tabular}
\end{table*}

What's more, to verify one of the strengths of our algorithm. That is, a subset of sample points is selected and used to train Brenier's height representation, while the remaining samples are directly predicted using the trained model. Tab. \ref{table:acc} displays the classification accuracy of \textbf{OT-Net} in domain adaptation. The results show that the method can predict the out-of-sample height vectors to directly obtain the entire height vector and OT map, and the algorithm is effective when applied to the domain adaptation task.

\section{Conclusion}
In this paper, a new neural network-based algorithm was presented to solve the OT map, which is a reusable optimal transport solver based on convex geometry. To be specific, when new samples are joined, it can directly utilize the learned  Brenier's height representation to calculate the height vector of out-of-sample, which can avoid recalculating or retraining the whole OT map. This greatly improves the computational effectiveness and reusability of the OT map. Moreover, we analyzed the error bound of the algorithm. Extensive experiments on both synthetic and real data demonstrate that our algorithm achieved comparable performance on generative models, color transfer, and domain adaptation.


\begin{appendices}


\section{}\label{secA1}
This section introduces the Encoder-Decoder architecture that is applied to the generative model and height representation network architecture of our algorithm. The autoencoder network structures are presented in Tab. \ref{table:encoder} and Tab. \ref{table:decoder}, and the height representation network structure can be found in Tab. \ref{table:heightrepresentation}, where $P_{data}$ and $P_{latent}$ represent the data distribution and the latent coding distribution, respectively. The Encoder-Decoder architecture was trained using the Adam algorithm with mini-batches of size 512, and learning rates of 2e-4, 1e-4, and 2e-5 in FASHION-MNIST \cite{Fashion}, Cifar-10 \cite{cifar10}, and CelebA \cite{celebA}, respectively. The height representation network was also trained using Adam with mini-batches of size 512, and learning rates of 0.004, 0.005, and 0.005 in FASHION-MNIST, Cifar-10, and CelebA, respectively.

\begin{table*}[htb!] 
\begin{center}
\begin{minipage}{\textwidth}
\caption{The Encoder architecture for CelebA, FASHION-MNIST and Cifar-10.}\label{table:encoder}
\scalebox{0.54}{
\begin{tabular}{lccccccccccc} 
\toprule%
& Input $x\sim P_{data}$ & \multicolumn{3}{c}{kernel size} & stride & padding & BN & activation & \multicolumn{3}{c}{number of outputs} \\
\midrule
&    & \multicolumn{3}{c}{CelebA ~~ FASHION-MNIST ~~Cifar-10} & & & & & \multicolumn{3}{l}{CelebA ~~~~~ FASHION-MNIST ~~~~Cifar-10} \\ \cmidrule{3-5}\cmidrule{10-12}

Layer1  & Convolution & 4$\ast$4  ~~~~~~~~~~~~~~& 4$\ast$4 & ~~~~4$\ast$4 & 2  & 1  & Yes  & LeakyReLU  & 32$\ast$32$\ast$dim  &  14$\ast$14$\ast$dim  & ~~~~16$\ast$16$\ast$dim \\
Layer2 & Convolution  & 4$\ast$4  ~~~~~~~~~~~~~~& 4$\ast$4 & ~~~~4$\ast$4 & 2  & 1  & Yes  & LeakyReLU  & 16$\ast$16$\ast$dim$\ast$2  & 7$\ast$7$\ast$dim$\ast$2 & ~~~~8$\ast$8$\ast$dim$\ast$2\\
Layer3 & Convolution  & 4$\ast$4  ~~~~~~~~~~~~~~& 3$\ast$3 & ~~~~4$\ast$4 & 2  & 1  & Yes  & LeakyReLU  & 8$\ast$8$\ast$dim$\ast$4  &  4$\ast$4$\ast$dim$\ast$4  &~~~~4$\ast$4$\ast$dim$\ast$4\\
Layer4 & Convolution  & 4$\ast$4  ~~~~~~~~~~~~~~& 4$\ast$4 & ~~~~4$\ast$4 & 2  & 1  & Yes  & LeakyReLU  & 4$\ast$4$\ast$dim$\ast$8  & 2$\ast$2$\ast$dim$\ast$8& ~~~~2$\ast$2$\ast$dim$\ast$8\\
Layer5 & Convolution  & 4$\ast$4  ~~~~~~~~~~~~~~& 2$\ast$2 & ~~~~2$\ast$2 & 1  & --  & --  & --        & 1$\ast$1$\ast$100  & 1$\ast$1$\ast$100 &~~~~1$\ast$1$\ast$100\\
\bottomrule
\end{tabular}
}
\end{minipage}
\end{center}
\end{table*}

\begin{table*}[htb!]  
\begin{center}
\begin{minipage}{\textwidth}
\caption{The Decoder architecture for CelebA, FASHION-MNIST and Cifar-10.}\label{table:decoder}
\scalebox{0.52}{
\begin{tabular}{lccccccccccc} 
\toprule%
& Input $y\sim P_{latent}$ & \multicolumn{3}{c}{kernel size} & stride & padding & BN & activation & \multicolumn{3}{c}{number of outputs} \\
\midrule
&    & \multicolumn{3}{c}{CelebA ~~ FASHION-MNIST ~~Cifar-10} & & & & & \multicolumn{3}{l}{CelebA ~~~~~~FASHION-MNIST ~~~~ Cifar-10} \\ \cmidrule{3-5}\cmidrule{10-12}

Layer1  & Transposed Convolution & 4$\ast$4  ~~~~~~~~~~~~~~& 2$\ast$2 & ~~~~~~~~2$\ast$2  & 1  & --  & --  & --    & 4$\ast$4$\ast$dim$\ast$8  & 2$\ast$2$\ast$dim$\ast$8 &~~2$\ast$2$\ast$dim$\ast$8\\
Layer2 & Transposed Convolution  & 4$\ast$4  ~~~~~~~~~~~~~~& 4$\ast$4 & ~~~~~~~~4$\ast$4  & 2  & 1  & Yes  & ReLU  & 8$\ast$8$\ast$dim$\ast$4  & 4$\ast$4$\ast$dim$\ast$4 &~~4$\ast$4$\ast$dim$\ast$4\\
Layer3 & Transposed Convolution  & 4$\ast$4  ~~~~~~~~~~~~~~& 3$\ast$3 & ~~~~~~~~4$\ast$4  & 2  & 1  & Yes  & ReLU  & 16$\ast$16$\ast$dim$\ast$2  & 7$\ast$7$\ast$dim$\ast$2 &~~8$\ast$8$\ast$dim$\ast$2\\
Layer4 & Transposed Convolution  & 4$\ast$4  ~~~~~~~~~~~~~~& 4$\ast$4 & ~~~~~~~~4$\ast$4  & 2  & 1  & Yes  & ReLU  & 32$\ast$32$\ast$dim  & 14$\ast$14$\ast$dim & ~~16$\ast$16$\ast$dim\\
Layer5 & Transposed Convolution  & 4$\ast$4  ~~~~~~~~~~~~~~& 4$\ast$4 &~~~~~~~~4$\ast$4  & 2  & 1  & --   & Tanh        & 64$\ast$64$\ast$3  & 28$\ast$28$\ast$3 & ~~32$\ast$32$\ast$3\\
\bottomrule
\end{tabular}
}
\end{minipage}
\end{center}
\end{table*}

\begin{table*}[htb!]  
\begin{center}
\begin{minipage}{\textwidth}
\caption{The height representation architecture for \textbf{OT-Net}.}\label{table:heightrepresentation}
\scalebox{0.71}{
\begin{tabular}{lcccc} 
\toprule%
& Input $y\sim P_{latent}$  & BN & activation &  number of outputs  \\
\midrule
& Nums$\ast$100 (Nums indicates the numbers of latent code features.) \\
\hline
Layer1  & Linear(100, 512) & Yes  & ReLU   & Nums$\ast$512 \\
Layer2 & Linear(512, 512)   & Yes  & ReLU  & Nums$\ast$512 \\
Layer3 & Linear(512, 512)  & Yes  & ReLU  & Nums$\ast$512 \\
Layer4 & Linear(512, 1)  & --  & --  & Nums$\ast$1 \\
\bottomrule 
\end{tabular}
}
\end{minipage}
\end{center}
\end{table*}



\end{appendices}


\section*{Declarations}

\bmhead{Author Contributions}
Zezeng Li provided original ideas and code implementation of the proposed algorithm. Shenghao Li was responsible for most of the experimental validation and manuscript writing. Lianbao Jin, Na Lei, and Zhongxuan Luo provided constructive ideas for theoretical derivation and experimental setup. All authors participated in the writing of the manuscript, and read and approved the final manuscript. 

\bmhead{Funding}
This research was supported by the National Key R$\&$D Program of China (2021YFA1003003),  and the National Natural Science Foundation of China under Grant (61936002, T2225012). 
\bmhead{Availability of data and material} 
The data/reanalysis that supports the findings of this study are publicly available online at \burl{http://yann.lecun.com/exdb/mnist/}, and \burl{https://github.com/zalandoresearch/fashion-mnist}, and 
\burl{http://www.cs.toronto.edu/kriz/cifar.html}, and
\burl{http://mmlab.ie.cuhk.edu.hk/projects/CelebA.html}.

\bmhead{Conflicts of interest/Competing interests}
The authors declare that they have no conflict of interest.
\bmhead{Ethics approval and Consent to participate} 
The authors declare that this research did not require Ethics
approval or Consent to participate since it does not concern human participants or human or animal datasets.

\bmhead{Consent for publication} The authors of this manuscript consent to its publication.

\bmhead{Code availability} 
The code can be obtained by contacting Shenghao Li and Zezeng Li.

\bibliography{otnet2}

\begin{thebibliography}{10}
\expandafter\ifx\csname url\endcsname\relax
  \def\url#1{\burl{#1}}\fi
\expandafter\ifx\csname urlprefix\endcsname\relax\def\urlprefix{URL }\fi
\providecommand{\bibinfo}[2]{#2}
\providecommand{\eprint}[2][]{\url{#2}}
\providecommand{\doi}[1]{\url{https://doi.org/#1}}
\bibcommenthead

\bibitem{seguy2018}
\bibinfo{author}{Seguy, V.} \emph{et~al.}
\newblock \bibinfo{title}{Large-scale optimal transport and mapping
  estimation}.
\newblock \emph{\bibinfo{journal}{ICLR 2018-International Conference on
  Learning Representations}} \bibinfo{pages}{1--15} (\bibinfo{year}{2018}).

\bibitem{chen2019gradual}
\bibinfo{author}{Chen, Y.} \emph{et~al.}
\newblock \bibinfo{title}{A gradual, semi-discrete approach to generative
  network training via explicit wasserstein minimization}.
\newblock \emph{\bibinfo{journal}{International Conference on Machine
  Learning}} \bibinfo{pages}{1071--1080} (\bibinfo{year}{2019}).

\bibitem{An2020AEOT}
\bibinfo{author}{An, D.} \emph{et~al.}
\newblock \bibinfo{title}{Ae-ot: A new generative model based on extended
  semi-discrete optimal transport}.
\newblock \emph{\bibinfo{journal}{ICLR 2020}}  (\bibinfo{year}{2019}).

\bibitem{liu2019wganqc}
\bibinfo{author}{Liu, H.}, \bibinfo{author}{Gu, X.} \&
  \bibinfo{author}{Samaras, D.}
\newblock \bibinfo{title}{Wasserstein gan with quadratic transport cost}.
\newblock \emph{\bibinfo{journal}{Proceedings of the IEEE/CVF international
  conference on computer vision}} \bibinfo{pages}{4832--4841}
  (\bibinfo{year}{2019}).

\bibitem{Daniels2021}
\bibinfo{author}{Daniels, M.}, \bibinfo{author}{Maunu, T.} \&
  \bibinfo{author}{Hand, P.}
\newblock \bibinfo{title}{Score-based generative neural networks for
  large-scale optimal transport}.
\newblock \emph{\bibinfo{journal}{Advances in neural information processing
  systems}} \textbf{\bibinfo{volume}{34}}, \bibinfo{pages}{12955--12965}
  (\bibinfo{year}{2021}).

\bibitem{Rout2021}
\bibinfo{author}{Rout, L.}, \bibinfo{author}{Korotin, A.} \&
  \bibinfo{author}{Burnaev, E.}
\newblock \bibinfo{title}{Generative modeling with optimal transport maps}.
\newblock \emph{\bibinfo{journal}{arXiv preprint arXiv:2110.02999}}
  (\bibinfo{year}{2021}).

\bibitem{gulrajani2017improvedgan}
\bibinfo{author}{Gulrajani, I.}, \bibinfo{author}{Ahmed, F.},
  \bibinfo{author}{Arjovsky, M.}, \bibinfo{author}{Dumoulin, V.} \&
  \bibinfo{author}{Courville, A.~C.}
\newblock \bibinfo{title}{Improved training of wasserstein gans}.
\newblock \emph{\bibinfo{journal}{Advances in neural information processing
  systems}} \textbf{\bibinfo{volume}{30}} (\bibinfo{year}{2017}).

\bibitem{courty2017joint}
\bibinfo{author}{Courty, N.}, \bibinfo{author}{Flamary, R.},
  \bibinfo{author}{Habrard, A.} \& \bibinfo{author}{Rakotomamonjy, A.}
\newblock \bibinfo{title}{Joint distribution optimal transportation for domain
  adaptation}.
\newblock \emph{\bibinfo{journal}{Advances in neural information processing
  systems}} \textbf{\bibinfo{volume}{30}} (\bibinfo{year}{2017}).

\bibitem{damodaran2018deepjdot}
\bibinfo{author}{Damodaran, B.~B.}, \bibinfo{author}{Kellenberger, B.},
  \bibinfo{author}{Flamary, R.}, \bibinfo{author}{Tuia, D.} \&
  \bibinfo{author}{Courty, N.}
\newblock \bibinfo{title}{Deepjdot: Deep joint distribution optimal transport
  for unsupervised domain adaptation}.
\newblock \emph{\bibinfo{journal}{Proceedings of the European conference on
  computer vision (ECCV)}} \bibinfo{pages}{447--463} (\bibinfo{year}{2018}).

\bibitem{wang2021}
\bibinfo{author}{Wang, W.}, \bibinfo{author}{Xu, H.}, \bibinfo{author}{Wang,
  G.}, \bibinfo{author}{Wang, W.} \& \bibinfo{author}{Carin, L.}
\newblock \bibinfo{title}{Zero-shot recognition via optimal transport}.
\newblock \emph{\bibinfo{journal}{Proceedings of the IEEE/CVF Winter Conference
  on Applications of Computer Vision}} \bibinfo{pages}{3471--3481}
  (\bibinfo{year}{2021}).

\bibitem{chang2022unified}
\bibinfo{author}{Chang, W.}, \bibinfo{author}{Shi, Y.}, \bibinfo{author}{Tuan,
  H.} \& \bibinfo{author}{Wang, J.}
\newblock \bibinfo{title}{Unified optimal transport framework for universal
  domain adaptation}.
\newblock \emph{\bibinfo{journal}{Advances in Neural Information Processing
  Systems}} \textbf{\bibinfo{volume}{35}}, \bibinfo{pages}{29512--29524}
  (\bibinfo{year}{2022}).

\bibitem{rakotomamonjy2022optimal}
\bibinfo{author}{Rakotomamonjy, A.} \emph{et~al.}
\newblock \bibinfo{title}{Optimal transport for conditional domain matching and
  label shift}.
\newblock \emph{\bibinfo{journal}{Machine Learning}} \bibinfo{pages}{1--20}
  (\bibinfo{year}{2022}).

\bibitem{chuang2023infoot}
\bibinfo{author}{Chuang, C.-Y.}, \bibinfo{author}{Jegelka, S.} \&
  \bibinfo{author}{Alvarez-Melis, D.}
\newblock \bibinfo{title}{Infoot: Information maximizing optimal transport}.
\newblock \emph{\bibinfo{journal}{International Conference on Machine
  Learning}} \bibinfo{pages}{6228--6242} (\bibinfo{year}{2023}).

\bibitem{tran2023unbalanced}
\bibinfo{author}{Tran, Q.~H.} \emph{et~al.}
\newblock \bibinfo{title}{Unbalanced co-optimal transport}.
\newblock \emph{\bibinfo{journal}{Proceedings of the AAAI Conference on
  Artificial Intelligence}} \textbf{\bibinfo{volume}{37}},
  \bibinfo{pages}{10006--10016} (\bibinfo{year}{2023}).

\bibitem{strossner2023low}
\bibinfo{author}{Str{\"o}ssner, C.} \& \bibinfo{author}{Kressner, D.}
\newblock \bibinfo{title}{Low-rank tensor approximations for solving
  multimarginal optimal transport problems}.
\newblock \emph{\bibinfo{journal}{SIAM Journal on Imaging Sciences}}
  \textbf{\bibinfo{volume}{16}}, \bibinfo{pages}{169--191}
  (\bibinfo{year}{2023}).

\bibitem{bonneel2019spot}
\bibinfo{author}{Bonneel, N.} \& \bibinfo{author}{Coeurjolly, D.}
\newblock \bibinfo{title}{Spot: sliced partial optimal transport}.
\newblock \emph{\bibinfo{journal}{ACM Transactions on Graphics (TOG)}}
  \textbf{\bibinfo{volume}{38}}, \bibinfo{pages}{1--13} (\bibinfo{year}{2019}).

\bibitem{Alvarez2018}
\bibinfo{author}{Alvarez-Melis, D.}, \bibinfo{author}{Jaakkola, T.} \&
  \bibinfo{author}{Jegelka, S.}
\newblock \bibinfo{title}{Structured optimal transport}.
\newblock \emph{\bibinfo{journal}{International conference on artificial
  intelligence and statistics}} \bibinfo{pages}{1771--1780}
  (\bibinfo{year}{2018}).

\bibitem{bonneel2016wasserstein}
\bibinfo{author}{Bonneel, N.}, \bibinfo{author}{Peyr{\'e}, G.} \&
  \bibinfo{author}{Cuturi, M.}
\newblock \bibinfo{title}{Wasserstein barycentric coordinates: histogram
  regression using optimal transport.}
\newblock \emph{\bibinfo{journal}{ACM Trans. Graph.}}
  \textbf{\bibinfo{volume}{35}}, \bibinfo{pages}{71--1} (\bibinfo{year}{2016}).

\bibitem{Ferradans2014}
\bibinfo{author}{Ferradans, S.}, \bibinfo{author}{Papadakis, N.},
  \bibinfo{author}{Peyr{\'e}, G.} \& \bibinfo{author}{Aujol, J.-F.}
\newblock \bibinfo{title}{Regularized discrete optimal transport}.
\newblock \emph{\bibinfo{journal}{SIAM Journal on Imaging Sciences}}
  \textbf{\bibinfo{volume}{7}}, \bibinfo{pages}{1853--1882}
  (\bibinfo{year}{2014}).

\bibitem{li2022real}
\bibinfo{author}{Li, Z.}, \bibinfo{author}{Lei, N.}, \bibinfo{author}{Shi, J.}
  \& \bibinfo{author}{Xue, H.}
\newblock \bibinfo{title}{Real-world super-resolution under the guidance of
  optimal transport}.
\newblock \emph{\bibinfo{journal}{Machine Vision and Applications}}
  \textbf{\bibinfo{volume}{33}}, \bibinfo{pages}{48} (\bibinfo{year}{2022}).

\bibitem{Gazdieva2022}
\bibinfo{author}{Gazdieva, M.}, \bibinfo{author}{Rout, L.},
  \bibinfo{author}{Korotin, A.}, \bibinfo{author}{Filippov, A.} \&
  \bibinfo{author}{Burnaev, E.}
\newblock \bibinfo{title}{Unpaired image super-resolution with optimal
  transport maps}.
\newblock \emph{\bibinfo{journal}{arXiv preprint arXiv:2202.01116}}
  (\bibinfo{year}{2022}).

\bibitem{li2022weakly}
\bibinfo{author}{Li, Z.}, \bibinfo{author}{Wang, W.}, \bibinfo{author}{Lei, N.}
  \& \bibinfo{author}{Wang, R.}
\newblock \bibinfo{title}{Weakly supervised point cloud upsampling via optimal
  transport}.
\newblock \emph{\bibinfo{journal}{ICASSP 2022-2022 IEEE International
  Conference on Acoustics, Speech and Signal Processing (ICASSP)}}
  \bibinfo{pages}{2564--2568} (\bibinfo{year}{2022}).

\bibitem{golla2020temporal}
\bibinfo{author}{Golla, T.}, \bibinfo{author}{Kneiphof, T.},
  \bibinfo{author}{Kuhlmann, H.}, \bibinfo{author}{Weinmann, M.} \&
  \bibinfo{author}{Klein, R.}
\newblock \bibinfo{title}{Temporal upsampling of point cloud sequences by
  optimal transport for plant growth visualization}.
\newblock \emph{\bibinfo{journal}{Computer Graphics Forum}}
  \textbf{\bibinfo{volume}{39}}, \bibinfo{pages}{167--179}
  (\bibinfo{year}{2020}).

\bibitem{cuturi2013sinkhorn}
\bibinfo{author}{Cuturi, M.}
\newblock \bibinfo{title}{Sinkhorn distances: Lightspeed computation of optimal
  transport}.
\newblock \emph{\bibinfo{journal}{Advances in neural information processing
  systems}} \textbf{\bibinfo{volume}{26}} (\bibinfo{year}{2013}).

\bibitem{bjd2015}
\bibinfo{author}{Benamou, J.-D.}, \bibinfo{author}{Carlier, G.},
  \bibinfo{author}{Cuturi, M.}, \bibinfo{author}{Nenna, L.} \&
  \bibinfo{author}{Peyr{\'e}, G.}
\newblock \bibinfo{title}{Iterative bregman projections for regularized
  transportation problems}.
\newblock \emph{\bibinfo{journal}{SIAM Journal on Scientific Computing}}
  \textbf{\bibinfo{volume}{37}}, \bibinfo{pages}{A1111--A1138}
  (\bibinfo{year}{2015}).

\bibitem{dpa2018}
\bibinfo{author}{Dvurechensky, P.}, \bibinfo{author}{Gasnikov, A.} \&
  \bibinfo{author}{Kroshnin, A.}
\newblock \bibinfo{title}{Computational optimal transport: Complexity by
  accelerated gradient descent is better than by sinkhorn’s algorithm}.
\newblock \emph{\bibinfo{journal}{International conference on machine
  learning}} \bibinfo{pages}{1367--1376} (\bibinfo{year}{2018}).

\bibitem{xie2020}
\bibinfo{author}{Xie, Y.}, \bibinfo{author}{Wang, X.}, \bibinfo{author}{Wang,
  R.} \& \bibinfo{author}{Zha, H.}
\newblock \bibinfo{title}{A fast proximal point method for computing exact
  wasserstein distance}.
\newblock \emph{\bibinfo{journal}{Uncertainty in artificial intelligence}}
  \bibinfo{pages}{433--453} (\bibinfo{year}{2020}).

\bibitem{anaaai2022}
\bibinfo{author}{An, D.}, \bibinfo{author}{Lei, N.}, \bibinfo{author}{Xu, X.}
  \& \bibinfo{author}{Gu, X.}
\newblock \bibinfo{title}{Efficient optimal transport algorithm by accelerated
  gradient descent}.
\newblock \emph{\bibinfo{journal}{Proceedings of the AAAI Conference on
  Artificial Intelligence}} \textbf{\bibinfo{volume}{36}},
  \bibinfo{pages}{10119--10128} (\bibinfo{year}{2022}).

\bibitem{Makkuva2020}
\bibinfo{author}{Makkuva, A.}, \bibinfo{author}{Taghvaei, A.},
  \bibinfo{author}{Oh, S.} \& \bibinfo{author}{Lee, J.}
\newblock \bibinfo{title}{Optimal transport mapping via input convex neural
  networks}.
\newblock \emph{\bibinfo{journal}{International Conference on Machine
  Learning}} \bibinfo{pages}{6672--6681} (\bibinfo{year}{2020}).

\bibitem{Fan2021}
\bibinfo{author}{Fan, J.}, \bibinfo{author}{Liu, S.}, \bibinfo{author}{Ma, S.},
  \bibinfo{author}{Chen, Y.} \& \bibinfo{author}{Zhou, H.}
\newblock \bibinfo{title}{Scalable computation of monge maps with general
  costs}.
\newblock \emph{\bibinfo{journal}{arXiv preprint arXiv:2106.03812}}
  \bibinfo{pages}{4} (\bibinfo{year}{2021}).

\bibitem{Korotin2022}
\bibinfo{author}{Korotin, A.}, \bibinfo{author}{Selikhanovych, D.} \&
  \bibinfo{author}{Burnaev, E.}
\newblock \bibinfo{title}{Neural optimal transport}.
\newblock \emph{\bibinfo{journal}{arXiv preprint arXiv:2201.12220}}
  (\bibinfo{year}{2022}).

\bibitem{Asadulaev2022}
\bibinfo{author}{Asadulaev, A.}, \bibinfo{author}{Korotin, A.},
  \bibinfo{author}{Egiazarian, V.} \& \bibinfo{author}{Burnaev, E.}
\newblock \bibinfo{title}{Neural optimal transport with general cost
  functionals}.
\newblock \emph{\bibinfo{journal}{arXiv preprint arXiv:2205.15403}}
  (\bibinfo{year}{2022}).

\bibitem{lei2019geometric}
\bibinfo{author}{Lei, N.} \emph{et~al.}
\newblock \bibinfo{title}{A geometric understanding of deep learning}.
\newblock \emph{\bibinfo{journal}{Engineering}} \textbf{\bibinfo{volume}{6}},
  \bibinfo{pages}{361--374} (\bibinfo{year}{2020}).

\bibitem{gu2016minkowski}
\bibinfo{author}{Gu, X.}, \bibinfo{author}{Luo, F.}, \bibinfo{author}{Sun, J.}
  \& \bibinfo{author}{Yau, S.-T.}
\newblock \bibinfo{title}{Variational principles for minkowski type problems,
  discrete optimal transport, and discrete monge--amp{\`e}re equations}.
\newblock \emph{\bibinfo{journal}{Asian Journal of Mathematics}}
  \textbf{\bibinfo{volume}{20}}, \bibinfo{pages}{383--398}
  (\bibinfo{year}{2016}).

\bibitem{Brenier1991}
\bibinfo{author}{Brenier, Y.}
\newblock \bibinfo{title}{Polar factorization and monotone rearrangement of
  vector-valued functions}.
\newblock \emph{\bibinfo{journal}{Communications on pure and applied
  mathematics}} \textbf{\bibinfo{volume}{44}}, \bibinfo{pages}{375--417}
  (\bibinfo{year}{1991}).

\bibitem{monge1781memoire}
\bibinfo{author}{Monge, G.}
\newblock \bibinfo{title}{M{\'e}moire sur la th{\'e}orie des d{\'e}blais et des
  remblais}.
\newblock \emph{\bibinfo{journal}{Histoire de l'Acad{\'e}mie Royale des
  Sciences de Paris}}  (\bibinfo{year}{1781}).

\bibitem{kantorovich1942transfer}
\bibinfo{author}{Kantorovich, L.}
\newblock \bibinfo{title}{On the transfer of masses (in russian)}.
\newblock \emph{\bibinfo{journal}{Doklady Akademii Nauk}}
  \textbf{\bibinfo{volume}{37}}, \bibinfo{pages}{227} (\bibinfo{year}{1942}).

\bibitem{Petzka2017}
\bibinfo{author}{Petzka, H.}, \bibinfo{author}{Fischer, A.} \&
  \bibinfo{author}{Lukovnicov, D.}
\newblock \bibinfo{title}{On the regularization of wasserstein gans}.
\newblock \emph{\bibinfo{journal}{arXiv preprint arXiv:1709.08894}}
  (\bibinfo{year}{2017}).

\bibitem{Sanjabi2018}
\bibinfo{author}{Sanjabi, M.}, \bibinfo{author}{Ba, J.},
  \bibinfo{author}{Razaviyayn, M.} \& \bibinfo{author}{Lee, J.~D.}
\newblock \bibinfo{title}{On the convergence and robustness of training gans
  with regularized optimal transport}.
\newblock \emph{\bibinfo{journal}{Advances in Neural Information Processing
  Systems}} \textbf{\bibinfo{volume}{31}} (\bibinfo{year}{2018}).

\bibitem{chen2017}
\bibinfo{author}{Chen, S.} \& \bibinfo{author}{Figalli, A.}
\newblock \bibinfo{title}{Partial w2, p regularity for optimal transport maps}.
\newblock \emph{\bibinfo{journal}{Journal of Functional Analysis}}
  \textbf{\bibinfo{volume}{272}}, \bibinfo{pages}{4588--4605}
  (\bibinfo{year}{2017}).

\bibitem{brenier1991polar}
\bibinfo{author}{Brenier, Y.}
\newblock \bibinfo{title}{Polar factorization and monotone rearrangement of
  vector-valued functions}.
\newblock \emph{\bibinfo{journal}{Communications on pure and applied
  mathematics}} \textbf{\bibinfo{volume}{44}}, \bibinfo{pages}{375--417}
  (\bibinfo{year}{1991}).

\bibitem{alexandrov2005convex}
\bibinfo{author}{Alexandrov, A.~D.}
\newblock \emph{\bibinfo{title}{Convex polyhedra}} Vol. \bibinfo{volume}{109}
  (\bibinfo{publisher}{Springer}, \bibinfo{year}{2005}).

\bibitem{adam2014}
\bibinfo{author}{Kingma, D.~P.} \& \bibinfo{author}{Ba, J.}
\newblock \bibinfo{title}{Adam: A method for stochastic optimization}.
\newblock \emph{\bibinfo{journal}{arXiv preprint arXiv:1412.6980}}
  (\bibinfo{year}{2014}).

\bibitem{MC1949}
\bibinfo{author}{Metropolis, N.} \& \bibinfo{author}{Ulam, S.}
\newblock \bibinfo{title}{The monte carlo method}.
\newblock \emph{\bibinfo{journal}{Journal of the American statistical
  association}} \textbf{\bibinfo{volume}{44}}, \bibinfo{pages}{335--341}
  (\bibinfo{year}{1949}).

\bibitem{MNIST}
\bibinfo{author}{LeCun, Y.}, \bibinfo{author}{Bottou, L.},
  \bibinfo{author}{Bengio, Y.} \& \bibinfo{author}{Haffner, P.}
\newblock \bibinfo{title}{Gradient-based learning applied to document
  recognition}.
\newblock \emph{\bibinfo{journal}{Proceedings of the IEEE}}
  \textbf{\bibinfo{volume}{86}}, \bibinfo{pages}{2278--2324}
  (\bibinfo{year}{1998}).

\bibitem{Fashion}
\bibinfo{author}{Xiao, H.}, \bibinfo{author}{Rasul, K.} \&
  \bibinfo{author}{Vollgraf, R.}
\newblock \bibinfo{title}{Fashion-mnist: a novel image dataset for benchmarking
  machine learning algorithms}.
\newblock \emph{\bibinfo{journal}{arXiv preprint arXiv:1708.07747}}
  (\bibinfo{year}{2017}).

\bibitem{cifar10}
\bibinfo{author}{Krizhevsky, A.}, \bibinfo{author}{Hinton, G.} \emph{et~al.}
\newblock \bibinfo{title}{Learning multiple layers of features from tiny
  images}.
\newblock \emph{\bibinfo{journal}{Master's thesis, University of Tront}}
  (\bibinfo{year}{2009}).

\bibitem{celebA}
\bibinfo{author}{Zhang, Z.}, \bibinfo{author}{Luo, P.}, \bibinfo{author}{Loy,
  C.~C.} \& \bibinfo{author}{Tang, X.}
\newblock \bibinfo{title}{From facial expression recognition to interpersonal
  relation prediction}.
\newblock \emph{\bibinfo{journal}{International Journal of Computer Vision}}
  \textbf{\bibinfo{volume}{126}}, \bibinfo{pages}{550--569}
  (\bibinfo{year}{2018}).

\bibitem{lin2018pacgan}
\bibinfo{author}{Lin, Z.}, \bibinfo{author}{Khetan, A.},
  \bibinfo{author}{Fanti, G.} \& \bibinfo{author}{Oh, S.}
\newblock \bibinfo{title}{Pacgan: The power of two samples in generative
  adversarial networks}.
\newblock \emph{\bibinfo{journal}{Advances in neural information processing
  systems}} \textbf{\bibinfo{volume}{31}} (\bibinfo{year}{2018}).

\bibitem{veegan}
\bibinfo{author}{Srivastava, A.}, \bibinfo{author}{Valkov, L.},
  \bibinfo{author}{Russell, C.}, \bibinfo{author}{Gutmann, M.~U.} \&
  \bibinfo{author}{Sutton, C.}
\newblock \bibinfo{title}{Veegan: Reducing mode collapse in gans using implicit
  variational learning}.
\newblock \emph{\bibinfo{journal}{Advances in neural information processing
  systems}} \textbf{\bibinfo{volume}{30}} (\bibinfo{year}{2017}).

\bibitem{Ugan2016}
\bibinfo{author}{Metz, L.}, \bibinfo{author}{Poole, B.}, \bibinfo{author}{Pfau,
  D.} \& \bibinfo{author}{Sohl-Dickstein, J.}
\newblock \bibinfo{title}{Unrolled generative adversarial networks}.
\newblock \emph{\bibinfo{journal}{arXiv preprint arXiv:1611.02163}}
  (\bibinfo{year}{2016}).

\bibitem{ALI2016}
\bibinfo{author}{Dumoulin, V.} \emph{et~al.}
\newblock \bibinfo{title}{Adversarially learned inference}.
\newblock \emph{\bibinfo{journal}{arXiv preprint arXiv:1606.00704}}
  (\bibinfo{year}{2016}).

\bibitem{MD2016}
\bibinfo{author}{Salimans, T.} \emph{et~al.}
\newblock \bibinfo{title}{Improved techniques for training gans}.
\newblock \emph{\bibinfo{journal}{Advances in neural information processing
  systems}} \textbf{\bibinfo{volume}{29}} (\bibinfo{year}{2016}).

\bibitem{heusel2017gansfid}
\bibinfo{author}{Heusel, M.}, \bibinfo{author}{Ramsauer, H.},
  \bibinfo{author}{Unterthiner, T.}, \bibinfo{author}{Nessler, B.} \&
  \bibinfo{author}{Hochreiter, S.}
\newblock \bibinfo{title}{Gans trained by a two time-scale update rule converge
  to a local nash equilibrium}.
\newblock \emph{\bibinfo{journal}{Advances in neural information processing
  systems}} \textbf{\bibinfo{volume}{30}} (\bibinfo{year}{2017}).

\bibitem{mmns2018}
\bibinfo{author}{Fedus, W.} \emph{et~al.}
\newblock \bibinfo{title}{Many paths to equilibrium: Gans do not need to
  decrease a divergence at every step}.
\newblock \emph{\bibinfo{journal}{arXiv preprint arXiv:1710.08446}}
  (\bibinfo{year}{2017}).

\bibitem{ls2017}
\bibinfo{author}{Mao, X.} \emph{et~al.}
\newblock \bibinfo{title}{Least squares generative adversarial networks}.
\newblock \emph{\bibinfo{journal}{Proceedings of the IEEE international
  conference on computer vision}} \bibinfo{pages}{2794--2802}
  (\bibinfo{year}{2017}).

\bibitem{wgan2017}
\bibinfo{author}{Arjovsky, M.}, \bibinfo{author}{Chintala, S.} \&
  \bibinfo{author}{Bottou, L.}
\newblock \bibinfo{title}{Wasserstein gan} (\bibinfo{year}{2017}).
\newblock \eprint{1701.07875}.

\bibitem{be2017}
\bibinfo{author}{Berthelot, D.}, \bibinfo{author}{Schumm, T.} \&
  \bibinfo{author}{Metz, L.}
\newblock \bibinfo{title}{Began: Boundary equilibrium generative adversarial
  networks}.
\newblock \emph{\bibinfo{journal}{arXiv preprint arXiv:1703.10717}}
  (\bibinfo{year}{2017}).

\bibitem{VAE2013}
\bibinfo{author}{Kingma, D.~P.} \& \bibinfo{author}{Welling, M.}
\newblock \bibinfo{title}{Auto-encoding variational bayes}.
\newblock \emph{\bibinfo{journal}{arXiv preprint arXiv:1312.6114}}
  (\bibinfo{year}{2013}).

\bibitem{GLO2017}
\bibinfo{author}{Bojanowski, P.}, \bibinfo{author}{Joulin, A.},
  \bibinfo{author}{Lopez-Paz, D.} \& \bibinfo{author}{Szlam, A.}
\newblock \bibinfo{title}{Optimizing the latent space of generative networks}.
\newblock \emph{\bibinfo{journal}{arXiv preprint arXiv:1707.05776}}
  (\bibinfo{year}{2017}).

\bibitem{hoshen2018non}
\bibinfo{author}{Hoshen, Y.}, \bibinfo{author}{Li, K.} \&
  \bibinfo{author}{Malik, J.}
\newblock \bibinfo{title}{Non-adversarial image synthesis with generative
  latent nearest neighbors}.
\newblock \emph{\bibinfo{journal}{Proceedings of the IEEE/CVF Conference on
  Computer Vision and Pattern Recognition}} \bibinfo{pages}{5811--5819}
  (\bibinfo{year}{2019}).

\bibitem{lucic2018gans}
\bibinfo{author}{Lucic, M.}, \bibinfo{author}{Kurach, K.},
  \bibinfo{author}{Michalski, M.} \emph{et~al.}
\newblock \bibinfo{title}{Are gans created equal? a large-scale study}.
\newblock \emph{\bibinfo{journal}{Advances in neural information processing
  systems}} \textbf{\bibinfo{volume}{31}} (\bibinfo{year}{2018}).

\bibitem{zhai2016generative}
\bibinfo{author}{Zhai, S.}, \bibinfo{author}{Cheng, Y.},
  \bibinfo{author}{Feris, R.} \& \bibinfo{author}{Zhang, Z.}
\newblock \bibinfo{title}{Generative adversarial networks as variational
  training of energy based models}.
\newblock \emph{\bibinfo{journal}{arXiv preprint arXiv:1611.01799}}
  (\bibinfo{year}{2016}).

\bibitem{abbasnejad2017bayesian}
\bibinfo{author}{Abbasnejad, M.~E.}, \bibinfo{author}{Shi, Q.},
  \bibinfo{author}{Abbasnejad, I.}, \bibinfo{author}{Hengel, A. v.~d.} \&
  \bibinfo{author}{Dick, A.}
\newblock \bibinfo{title}{Bayesian conditional generative adverserial
  networks}.
\newblock \emph{\bibinfo{journal}{arXiv preprint arXiv:1706.05477}}
  (\bibinfo{year}{2017}).

\bibitem{rosca2017variational}
\bibinfo{author}{Rosca, M.}, \bibinfo{author}{Lakshminarayanan, B.},
  \bibinfo{author}{Warde-Farley, D.} \& \bibinfo{author}{Mohamed, S.}
\newblock \bibinfo{title}{Variational approaches for auto-encoding generative
  adversarial networks}.
\newblock \emph{\bibinfo{journal}{arXiv preprint arXiv:1706.04987}}
  (\bibinfo{year}{2017}).

\bibitem{ns2011}
\bibinfo{author}{Bonneel, N.}, \bibinfo{author}{Van De~Panne, M.},
  \bibinfo{author}{Paris, S.} \& \bibinfo{author}{Heidrich, W.}
\newblock \bibinfo{title}{Displacement interpolation using lagrangian mass
  transport}.
\newblock \emph{\bibinfo{journal}{Proceedings of the 2011 SIGGRAPH Asia
  conference}} \bibinfo{pages}{1--12} (\bibinfo{year}{2011}).

\bibitem{pot2021}
\bibinfo{author}{Flamary, R.} \emph{et~al.}
\newblock \bibinfo{title}{Pot: Python optimal transport}.
\newblock \emph{\bibinfo{journal}{The Journal of Machine Learning Research}}
  \textbf{\bibinfo{volume}{22}}, \bibinfo{pages}{3571--3578}
  (\bibinfo{year}{2021}).

\bibitem{bonneel2011displacement}
\bibinfo{author}{Bonneel, N.}, \bibinfo{author}{Van De~Panne, M.},
  \bibinfo{author}{Paris, S.} \& \bibinfo{author}{Heidrich, W.}
\newblock \bibinfo{title}{Displacement interpolation using lagrangian mass
  transport}.
\newblock \emph{\bibinfo{journal}{Proceedings of the 2011 SIGGRAPH Asia
  conference}} \bibinfo{pages}{1--12} (\bibinfo{year}{2011}).

\bibitem{OTDA2016}
\bibinfo{author}{Flamary, R.}, \bibinfo{author}{Courty, N.},
  \bibinfo{author}{Tuia, D.} \& \bibinfo{author}{Rakotomamonjy, A.}
\newblock \bibinfo{title}{Optimal transport for domain adaptation}.
\newblock \emph{\bibinfo{journal}{IEEE Trans. Pattern Anal. Mach. Intell}}
  \textbf{\bibinfo{volume}{1}} (\bibinfo{year}{2016}).

\end{thebibliography}

\end{document}